\documentclass{article}

\usepackage{microtype}
\usepackage{graphicx}
\usepackage{subfigure}
\usepackage{booktabs} 

\usepackage{hyperref}

\usepackage[accepted]{icml2024}

\usepackage{amsmath}
\usepackage{amssymb}
\usepackage{mathtools}
\usepackage{amsthm}

\usepackage[capitalize,noabbrev]{cleveref}
\usepackage{multicol}
\usepackage{multirow}
\theoremstyle{plain}

\theoremstyle{definition}

\theoremstyle{remark}

\usepackage[textsize=tiny]{todonotes}

\icmltitlerunning{AD3: Implicit Action is the Key for World Models to Distinguish the Diverse Visual Distractors}

\begin{document}

\twocolumn[
\icmltitle{AD3: Implicit Action is the Key for World Models to Distinguish the Diverse Visual Distractors}



\icmlsetsymbol{equal}{*}

\begin{icmlauthorlist}
\icmlauthor{Yucen Wang}{equal,nju,lamda}
\icmlauthor{Shenghua Wan}{equal,nju,lamda}
\icmlauthor{Le Gan}{nju,lamda}
\icmlauthor{Shuai Feng}{bit}
\icmlauthor{De-Chuan Zhan}{nju,lamda}
\end{icmlauthorlist}

\icmlaffiliation{nju}{School of Artificial Intelligence, Nanjing University, China}
\icmlaffiliation{lamda}{National Key Laboratory for Novel Software Technology, Nanjing University, China}
\icmlaffiliation{bit}{School of Cyberspace Science and Technology, Beijing Institute of Technology}

\icmlcorrespondingauthor{De-Chuan Zhan}{zdc@nju.edu.cn}

\icmlkeywords{Machine Learning, ICML}

\vskip 0.3in
]



\printAffiliationsAndNotice{\icmlEqualContribution} 

\begin{abstract}
Model-based methods have significantly contributed to distinguishing task-irrelevant distractors for visual control. However, prior research has primarily focused on heterogeneous distractors like noisy background videos, leaving homogeneous distractors that closely resemble controllable agents largely unexplored, which poses significant challenges to existing methods. To tackle this problem, we propose Implicit Action Generator (IAG) to learn the implicit actions of visual distractors, and present a new algorithm named \emph{implicit \textbf{A}ction-informed \textbf{D}iverse visual \textbf{D}istractors \textbf{D}istinguisher} (AD3), that leverages the action inferred by IAG to train separated world models. Implicit actions effectively capture the behavior of background distractors, aiding in distinguishing the task-irrelevant components, and the agent can optimize the policy within the task-relevant state space. Our method achieves superior performance on various visual control tasks featuring both heterogeneous and homogeneous distractors. The indispensable role of implicit actions learned by IAG is also empirically validated.
\end{abstract}

\section{Introduction}
\label{sec: Introduction}

In recent years, there has been a surge in research on reinforcement learning with visual inputs  \cite{yarats2020image, yarats2021mastering, laskin2020curl, hafner2019dream, DBLP:conf/iclr/HafnerL0B21, hafner2023mastering}. This interest has arisen from the application of RL algorithms to real-world scenarios, where visual inputs are often filled with distractive elements unrelated to the task. Model-based methods have significantly contributed to distinguishing distractors and extracting task-relevant information for visual control, as demonstrated by approaches such as TIA \cite{DBLP:conf/icml/FuYAJ21} and Denoised MDP \cite{DBLP:conf/icml/0001D0IZT22}. However, prior research has primarily focused on eliminating heterogeneous distractors, such as noisy video backgrounds that exhibit entirely different visual semantics from the agent behavior. When confronted with homogeneous distractors that closely resemble the task-relevant controllable elements, previous methods often encounter substantial difficulties. For instance, when the visual observation contains an additional shifted agent that shares a similar morphological structure   with the controllable agent but cannot be directly 
manipulated \cite{DBLP:conf/iclr/BharadhwajBEL22}, the policy will get confused in determining which one to control.  

For model-based approaches primarily reliant on observation reconstruction such as Dreamer  \cite{hafner2019dream}, the inclusion of task-irrelevant information, like the background agent, presents considerable challenges for extracting task signals. TIA \cite{DBLP:conf/icml/FuYAJ21} seeks to disentangle the latent state into task signal and noise, through adversarial reward dissociation on the task-irrelevant state and distractor-only reconstruction. However, such objectives are sensitive, with optimization instability and inappropriate loss weights easily leading to the inversion of the two components. Moreover, the underlying assumption behind these heuristic loss objectives may not hold in environments featuring homogeneous distractors, where task-irrelevant information constitutes a relatively small proportion of the observation. Another category of distractor-eliminating methods is based on state factorization, as seen in Denoised MDP  \cite{DBLP:conf/icml/0001D0IZT22}, but the assumed transition structures of different factors are not enough to distinguish homogeneous distractors. Other model-free approaches  \cite{laskin2020curl, yarats2020image, DBLP:conf/iclr/0001MCGL21} and visual-based methods also struggle to guarantee effective distractor removal, especially on visually indistinguishable distractors. 

To a certain extent, such homogeneous visual distractors can more accurately reflect the essence of RL problems compared to heterogeneous ones: distinguishment can only be achieved by leveraging the inherent characteristics of the \textbf{control} problem itself to identify the semantics and dynamic changes of task-relevant and irrelevant components, rather than relying on other visual elements, as distractors are not visually distinguishable when they are homogeneous. In this context, the significance of actions becomes evident, which takes precedence over observations and rewards, specifically the implicit actions of distractors. Apparently, if we can obtain the actions executed by the shifted agent, the semantics of homogeneous distractor will be easily captured. We empirically validate in \cref{section: action_type} that near-optimal separation of the two components can be realized by utilizing agent actions and precise actions of the distractor to construct separated world models. Similar principle also holds for addressing heterogeneous distractors, where the dynamic changes within noisy backgrounds, such as natural videos, also exhibit semantic transitions. These transitions can also be encoded through the implicit actions "executed" by the noisy components.

Building upon these insights, we propose the Implicit-Action Block MDP (IABMDP), which operates under the assumption that the dynamics of task-relevant and irrelevant components are respectively conditioned on agent actions and implicit actions of existing distractors. In line with this assumption, we propose a practical method named \emph{implicit \textbf{A}ction-informed \textbf{D}iverse visual \textbf{D}istractors \textbf{D}istinguisher} (AD3). AD3 is designed to learn separated world models with a focus on utilizing the inferred implicit actions of distractors. Such inference can be achieved by the proposed Implicit Action Generator (IAG), which leverages cycle consistency  \cite{FICC23, dwibedi2019temporal, wang2019learning} to extract the semantics of task-irrelevant dynamic transitions by implicitly decoupling the impact of task-relevant agent actions. We also employ categorical variables to bottleneck the representation of the implicit action. Furthermore, we propose to use these implicit actions of distractors and agent actions to construct separated world models for task-relevant and irrelevant components, respectively, through standard variational inference, without the need for additional loss objectives. Policy learning is exclusively conducted within the task-relevant world model and latent state space.

To begin with, we evaluate our method on DeepMind Control Suite tasks  \cite{tassa2018deepmind} in the context of heterogeneous and homogeneous distractors. AD3 consistently performs well on various tasks with either of the  distractors. Furthermore, empirical study exhibits the irreplaceable significance of the implicit action learned by IAG for distinguishing task-irrelevant components, as well as the interpretable semantics of the implicit action. From those sides, our proposed method is a general approach to identifying and eliminating various types of visual distractors, and the notion of implicit action inference can be widely used.

Our contributions can be summarized as follows: (i) We propose a new module, Implicit Action Generator (IAG), that can generate implicit actions of existing distractors and be plugged into any model-based visual RL methods that aim to model the task-irrelevant dynamics. (ii) We propose a new method, Implicit Action-Informed Diverse Visual Distractors Distinguisher (AD3), which can distinguish diverse visual distractors including both heterogeneous and underexplored homogeneous ones. (iii) Our method achieves superior performance on various visual control tasks with different distractors.

\section{Related Work}
\label{sec: Related_work}

\textbf{Visual Reinforcement Learning with Noisy Observations.}\label{sec:rl with noise} 
Many recent studies in RL explored ways to enhance performance in visual environments with noise or distractors, which can be categorized into two groups: reconstruction-free and reconstruction-based. For instance, DBC \cite{DBLP:conf/iclr/0001MCGL21} learns a compact latent state by bisimulation metric to filter out distractors in the environment. InfoPower \cite{DBLP:conf/iclr/BharadhwajBEL22} combines a variational empowerment term into the state-space model to capture task-relevant features at first. These works substitute the reconstruction's functionality with other designs. For reconstruction-based methods, Iso-Dream \cite{pan2022iso} builds decoupled world models on isolated environment states based on controllability and inverse dynamics prediction. Denoised MDP \cite{DBLP:conf/icml/0001D0IZT22} decomposites the visual observation into four parts by action and reward, and constructs the corresponding models. The most similar method to us is TIA  \cite{DBLP:conf/icml/FuYAJ21}, which also learns separated world models for capturing the task and distractor features. However, TIA does not learn an implicit action to induce the task-irrelevant dynamics and directly relies on agent actions, which substantially differs from our method.

\textbf{Learning Latent Actions in RL.} Several prior works learn latent actions to enhance policy performance and sample efficiency. For instance, PG-RA \cite{PGRA19} learns action representations that enhance generalization across large action sets. LASER \cite{LASER21} disentangles raw actions into a latent space aligned with the task domain. These works infer latent actions solely from observations. Some other research focuses on learning representations for predefined actions. ILPO \cite{ILPO19} infers latent actions from expert observations and aligns them with real-world actions. FICC \cite{FICC23} and LAPO \cite{schmidt2024learning} both employ a two-phase training pipeline with a cycle consistency objective to learn latent actions. SWIM \cite{swim23rss} constructs a structured human-centric action space based on visual affordances, and TAP \cite{TAP23} uses a state-conditional VQ-VAE to learn low-dimensional latent action codes. Our method can be regarded as the combination of these two directions, which infer the underlying actions of distractors from observations with agent actions helping identify the task-related information.

\section{Preliminaries}
\label{sec: Preliminaries}

\textbf{Block MDP.} In RL problems with high-dimensional inputs such as images and videos, the real state space of the task is hidden, necessitating its estimation from the observation. Many prior works have made the Block MDP assumption  \cite{du2019provably} for this scenario. A Block MDP can be defined as a tuple of $\mathcal{M}=(\mathcal{O}, \mathcal{Z}, \mathcal{A}, \mathcal{T}, \mathcal{R}, \mathcal{U}, \mu_0)$ which respectively represents observations, latent states, actions, transition function, reward function, emission function, and the initial state distribution. Block MDP hypothesizes that each observation $o$ is uniquely generated by a latent state $z$ through the emission function $\mathcal{U}$, so that the observation contains enough information to decode the corresponding real state exclusively. Real-world pixel-based RL problems where observations contain rich semantics and information about the task can benefit from this assumption. 

\textbf{Implicit-Action Block MDP.}
To tackle pixel-based RL problems with visual distractors, we propose the Implicit-Action Block MDP (IABMDP). We assume that the background distractors have their own implicit actions to give rise to their visual changes and 
 dynamic transitions. This implicit distractor action can significantly aid in characterizing 
 the noise and task-irrelevant elements within the observation. IABMDP adds a set of distractor actions $\mathcal{A}^{-}$ to the 7-tuple of Block MDP, 
 and a latent state $z_t$ in IABMDP can be decoupled into task-relevant and -irrelevant components: $z_t=(z_t^+, z_t^-)$, whose dynamics are also decomposed and are respectively conditioned on agent actions $a_t$ and implicit actions $a_t^{-}$ from existing distractors:
\begin{equation}
\label{iabmdp_assump_1}
    \begin{aligned}
        &\text{Task-relevant MDP} & z_{t+1}^+ &\sim \mathcal{T}_+(\cdot \mid z_t^+, a_t) \\
        & &r_t & \sim\mathcal{R}(\cdot \mid z_t^+)\\
        &\text{Task-irrelevant MDP} & z_{t+1}^- &\sim \mathcal{T}_-(\cdot \mid z_t^-, a_t^-)
    \end{aligned}
\end{equation}
\begin{equation}
\label{iabmdp_assump_2}
    \mathcal{T}(z_{t+1}|z_t,a_t,a_t^-) = \mathcal{T}_+(z^+_{t+1} | z^+_t,a_t) \mathcal{T}_-(z^-_{t+1}|z_t^-,a_t^-)
\end{equation}
where $z_t^+$ represents the task-related latent state that can be controlled by the agent, while $z_t^-$ characterizes the task-irrelevant component with its distinct transition dynamics influenced solely by the implicit action $a_t^-$. IABMDP processes the two components independently, and the separate world model learning can thus be realized as long as we find $a_t^-$, the implicit action of distractors. The graphic model of IABMDP assumption is shown in \cref{IABMDP and IAG architecture}(a).

It is worth highlighting that our underlying assumption fundamentally differs from that of TiMDP \cite{DBLP:conf/icml/FuYAJ21}, EX-BMDP  \cite{efroni2021provable}, Iso-Dream \cite{pan2022iso} and AcT \cite{wan2023semail}. These approaches either employ agent actions to induce the task-irrelevant MDP or assume that distractor transitions are independent of any action. Our assumption deviates from prior methods primarily by explicitly introducing the concept of implicit distractor actions to extract task-irrelevant components. This distinctive feature sets our method apart from existing approaches. We present a detailed comparative analysis of these methodologies in \cref{appendix: assumptions decomposition}.

\textbf{Model-based Reinforcement Learning (MBRL).} 
MBRL is a paradigm that entails learning the environment's dynamics and reward function from experience. The model is then utilized to formulate action plans by exploring potential future states \cite{sutton2018reinforcement}, enabling sample-efficient learning as the agent can learn from the simulated environment instead of the real one \cite{moerland2023model, luo2022survey}. While initially proposed for state-based RL problems  \cite{sutton1990integrated, janner2019trust, yu2020mopo}, subsequent research has demonstrated its particular efficacy for visual inputs \cite{ha2018world}, since Visual MBRL can help obtain a low-dimensional surrogate environment, thereby reducing storage requirements and enhancing learning efficiency. The success of this approach is exemplified in Dreamer and its extensions \cite{hafner2019learning, hafner2019dream, DBLP:conf/iclr/HafnerL0B21, hafner2023mastering}, where a compact latent state space is learned by maximizing the Evidence Lower Bound (ELBO) using Recurrent State Space Model (RSSM) architecture. Our method employs Dreamer-style world model as a backbone.

\section{Methods}
\label{sec: method}

We employ the IABMDP assumption to model the RL process from visual observations containing complex distractors. Specifically, we construct separate MDPs for task-relevant and irrelevant components. In~\Cref{subsec: IAG}, we design the Implicit Action Generator (IAG) to infer possible actions for the task-irrelevant part. In~\Cref{subsec: ASWM}, we present our approach for constructing world models for both task-relevant and task-irrelevant components by leveraging agent actions and the inferred implicit actions, respectively. We obtain the ELBO of the variational objective and minimize the overall loss to update the separated world models. In~\Cref{subsec: PL}, we introduce how to train the policy in the task-relevant world model by imagination.

\subsection{Implicit Action Generator}
\label{subsec: IAG}

\begin{figure}[t]
\begin{center}
\centerline{\includegraphics[width=\columnwidth]{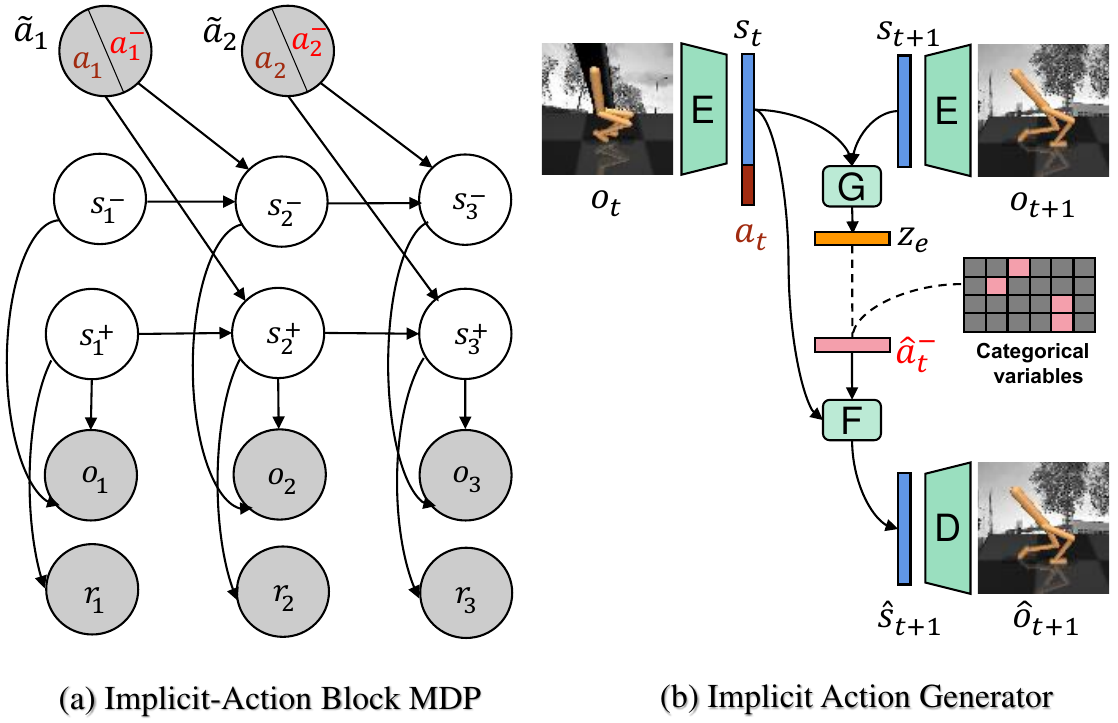}}
\caption{The IABMDP assumption and the architecture of IAG}
\label{IABMDP and IAG architecture}
\end{center}
\end{figure}

We propose two dynamic models to help identify the implicit actions of distractors: Task-relevant Action-conditioned Inverse Dynamics (TAID) and Forward Implicit Action-conditioned Dynamics (FIAD).
\begin{equation}\label{eq:IAG-model}
\begin{aligned}
    &\text{TAID: }& \hat{a}_t^- = G(s_t, a_t, s_{t+1})\\
    &\text{FIAD: }& \hat{s}_{t+1} = F(s_t, a_t, \hat{a}_t^-)
\end{aligned}
\end{equation}
Here, $\hat{a}_t^-$ is the estimated implicit action of the distractor. To help understand the intuition behind TAID and FIAD, we can reformulate them with the IABMDP assumption. They can be rewritten as 
\begin{equation}\label{eq:IAG-model-interpretation}
\begin{aligned}
    &\text{TAID: }& \hat{a}^-_t &=G([s_t^+;s_t^-], a_t, [s_{t+1}^+;s_{t+1}^-])
    \\ & & &= G(\underbrace{s^{+}_t, a_t, s^{+}_{t+1}}_{f_+},\underbrace{s^{-}_t, s^{-}_{t+1}}_{g_-}) 
    \\ 
    &\text{FIAD: }& \hat{s}_{t+1} &= F([s_t^+;s_t^-],[a_t;\hat{a}_t^-]) \\
    &&&  =F(\underbrace{s^{+}_t, a_t}_{f_+},\underbrace{s^{-}_t, \hat{a}^{-}_{t}}_{f_-}) 
    \\
    & & &= [\hat{s}_{t+1}^+;\hat{s}_{t+1}^-] 
\end{aligned}
\end{equation}
where $f$ and $g$ denote the forward and inverse dynamics, and subscripts $+$ and $-$ denote the task-relevant and irrelevant parts. TAID identifies task-relevant states $s_t^{+},s_{t+1}^+$ by implicit forward dynamic $f_+$ conditioned on the agent action $a_t$, resulting in that the irrelevant transition between $s^{-}_t$ and $s^{-}_{t+1}$ is only conditioned on the implicit action, i.e., the implicit action can only be inferred from task-irrelevant parts. TAID implicitly learns an inverse dynamic $g_-$ to infer $\hat{a}^-$. To maintain cycle consistency, we use FIAD to generate the prediction of the next state $\hat{s}_{t+1}$ consisting of both task-relevant and irrelevant parts by implicitly learned forward dynamics $f_+, f_-$. Notably, all the notations in the underbraces provide interpretations of our approach, but do not have a one-to-one correspondence to the practical modeling process. 

Obviously, TAID and FIAD can be modeled by an auto-encoder, where the encoder $G$ takes $(s_t, a_t)$ and $s_{t+1}$ as inputs to infer the implicit action $\hat{a}_t^-$ of the distractor, and the decoder $F$ is again fed in $(s_t, a_t)$ and $\hat{a}_t^-$ inferred by the encoder to generate the forward-prediction result $\hat{s}_{t+1}$ as an output. We only need the encoder to get the implicit action during inference.
 
 Various loss objectives based on consistency or reconstruction can be designed to learn the encoder and decoder of IAG. We maximize the cosine similarity of $\hat{s}_{t+1}$ and $s_{t+1}$ to ensure their closeness, enhancing the cycle consistency required by FIAD. We also propose a difference reconstruction term at pixel level with $(s_t, a_t)$ and the inferred $\hat{a}_t^-$ as inputs to force TAID to concentrate on the changes occurring in the environment. Note that both $a_t$ and $\hat{a}_t^-$ are involved in difference reconstruction, which urges $\hat{a}_t^-$ to focus on the changes that cannot be accomplished by $a_t$, since it is impossible for solely $a_t$ to reconstruct the difference of the whole observation between timesteps given the existence of distractors. Hence, the difference reconstruction loss objective shares a similar motivation with TAID. By optimizing these two objectives, we can implicitly extract $s_t^{-}$ and $s^{-}_{t+1}$ that can not be influenced by $a_t$, and enforce the learned $\hat{a}_t^-$ to realize the transition from $s_t^{-}$ to $s_{t+1}^{-}$. Moreover, a one-step image reconstruction loss objective is also needed to encode full information into the latent state $s_{t}$, which is beneficial for the previous two objectives. The insights behind these loss designs are widely used \cite{FICC23, schmidt2024learning, ILPO19}, which are found effective in learning representations in much previous research including FICC \cite{FICC23}. The objective in the Implicit-Action Generation process is as follows:
\begin{equation}
\label{eq:IAG}
\begin{aligned}
    \mathcal{L}_{\text{IAG}} = -\cos(\hat{s}_{t+1}, s_{t+1}) & - \ln p(o_{t+1} - o_t|s_t, a_t, \hat{a}_t^-) \\ & - \ln p(o_t|s_t)
\end{aligned}
\end{equation}

The above process may cause shortcuts that result in meaningless latent action space \cite{FICC23}. For instance, TAID may simply copy $s_{t+1}$ into the latent space and directly output it through FIAD with zero reconstruction loss, with the impact of $s_t, a_t$ and the dynamics in the environment totally ignored. 

To avoid shortcuts, FICC leverages vector quantization technique \cite{kohonen2001learning,van2017neural} which has the ability to reduce the amount of information that the latent space contains, and learn interpretable discrete representations. For the sake of stronger bottlenecking of the learned latent action space, we consider the categorical latent variables technique, which is employed in DreamerV2 \cite{DBLP:conf/iclr/HafnerL0B21, NEURIPS2022_a766f56d} and brings about marvelous performance boost. Compared to traditional quantization methods \cite{kohonen2001learning,van2017neural,2023qlae}, categorical (one-hot) coding can be more flexible since it implicitly builds different latent codebooks for different learning objectives. The learning process of the codebooks and the quantizing operation are inherently embedded in the categorical-variable architecture, and using one-hot codes in forward propagation equals to leveraging active bits in the code as indices for quantization. Moreover, the extreme sparsity of categorical variables significantly benefits the bottlenecking of the latent action space in IAG and enhances generalization \cite{DBLP:conf/iclr/HafnerL0B21}. More implementation details and discussions on using categorical variables as implicit action representations can be found in \cref{appendix: one-hot coding implementation} and \cref{appendix: categorical latents}.

To summarize, the aforementioned three design components, namely the conceptual  framework, loss objectives, and representation form, collectively constitute the complete structure of the Implicit Action Generator. IAG can infer a meaningful implicit action $\hat{a}^-$ representing the semantics of the distractor's dynamic transition, which is ready to be utilized to learn the task-irrelevant MDP. The architecture of IAG is shown in \cref{IABMDP and IAG architecture}(b).

\subsection{Action-conditioned Separated World Models} \label{subsec: ASWM}
Based on the IABMDP assumption, we can separately construct the task-related and task-irrelevant models by utilizing agent actions $a_t$ and implicit distractor actions $\hat{a}_t^-$ inferred by IAG, respectively. The loss objective for learning the separated world models can be derived by constructing the Evidence Lower Bound (ELBO) of the log-likelihood of the observed data predicted by the decomposed latent states, taking into account both agent actions and the implicit actions of distractors. The derivation of the loss function below is in \cref{appendix: deriv}.
\begin{equation}
\label{AD3ELBO}
\begin{split}
     & \mathcal{L}_{\mathcal{\widetilde{M}}}  =  \sum_{t=1}^T \Bigg[ \mathbb{E}_{p_{\psi^+} , p_{\psi^-}} \Big[ -\ln q_{\phi}\left(o_t \mid z_t^+, z_t^-\right) +  \\
    &  \mathbb{D}_{\text{KL}} \left[p_{\psi^+}\left(\cdot \mid o_{\leq t}, a_{<t}\right) \| q_{\theta^+}\left(\cdot \mid z^+_{t-1}, a_{t-1}\right)\right] + \\
     &  \mathbb{D}_{\text{KL}} \left[p_{\psi^-}\left(\cdot \mid o_{\leq t}, \hat{a}_{<t}^-\right) \| q_{\theta^-}\left(\cdot \mid z^-_{t-1}, \hat{a}^-_{t-1}\right)\right]\Big]\Bigg]
\end{split}
\end{equation}
 where $p_{\psi^+}$ and $p_{\psi^-}$ are two separated variational encoders to infer the task-related and task-irrelevant latent states $z_t^+$ and $z_t^-$ from historical data including observations and agent / distractor actions. $q_{\theta^+}(z^{+}_{t+1}|z^{+}_t,a_t)$ and $q_{\theta^-}(z^-_{t+1}|z^-_t, \hat{a}^-_{t})$ are two forward transition models for separately learning the dynamics of the agent and the distractor. Moreover, $q_{\phi}$ is the decoder function to jointly reconstruct the whole observation $o_t$ by task-relevant and -irrelevant states. The equation presented in \cref{AD3ELBO} is analogous to the objectives of two distinct Dreamer models plus cooperative image reconstruction from their respective latent states. The reward function also needs to be learned by task-relevant states, with the loss objective $-\ln q_{\omega}(r_t \mid z^+_{t})$.

With respect to the implementation of action-conditioned separated world models, we design two independent Recurrent State Space Models (RSSM)  \cite{hafner2019learning}, one incorporating agent actions and the other utilizing the inferred implicit actions as inputs. For cooperative image reconstruction, we follow the masking approach in TIA and utilize two independent observation decoders $q_{\phi^+}$ and $q_{\phi^-}$. 

\subsection{Policy Learning} \label{subsec: PL}
We train the policy only in the task-relevant latent state space by imagining rollouts through the learned reward function and task-relevant transition model. 
\begin{equation}
\label{eq:policy}
    \begin{split}
        &\text{Action model:}\; a_t\sim \pi\left(a_t|z^+_t\right)\\
        &\text{Value model:}\; v\left(z_t^+\right) \approx \mathbb{E}_{\pi, q_{\theta^{+}}}\Big[\sum_{k=t}^{H}\gamma^{k-t} q_{\omega} \left(z_k^+, a_k\right)\Big]\nonumber
    \end{split}
\end{equation}
\paragraph{Remarks.} The AD3 method involves iterative learning of IAG and separated world models as well as the policy. Current action inference results given by IAG are assigned to the trajectory data in the buffer every time IAG makes a training update, which can be utilized in learning the task-irrelevant branch of the separated world models. Gradients of IAG will not flow to world-model learning. The pseudo codes of the AD3 method are shown in \cref{alg:AD3}, and more implementation details can be found in \cref{appendix: implementation networks and hyperparameters}. Notably, the objective function in \cref{AD3ELBO} is a clear ELBO rigorously derived through standard variational inference without any other heuristic human bias injected. 

\begin{figure*}[t]
\begin{center}
\centerline{\includegraphics[width=\textwidth]{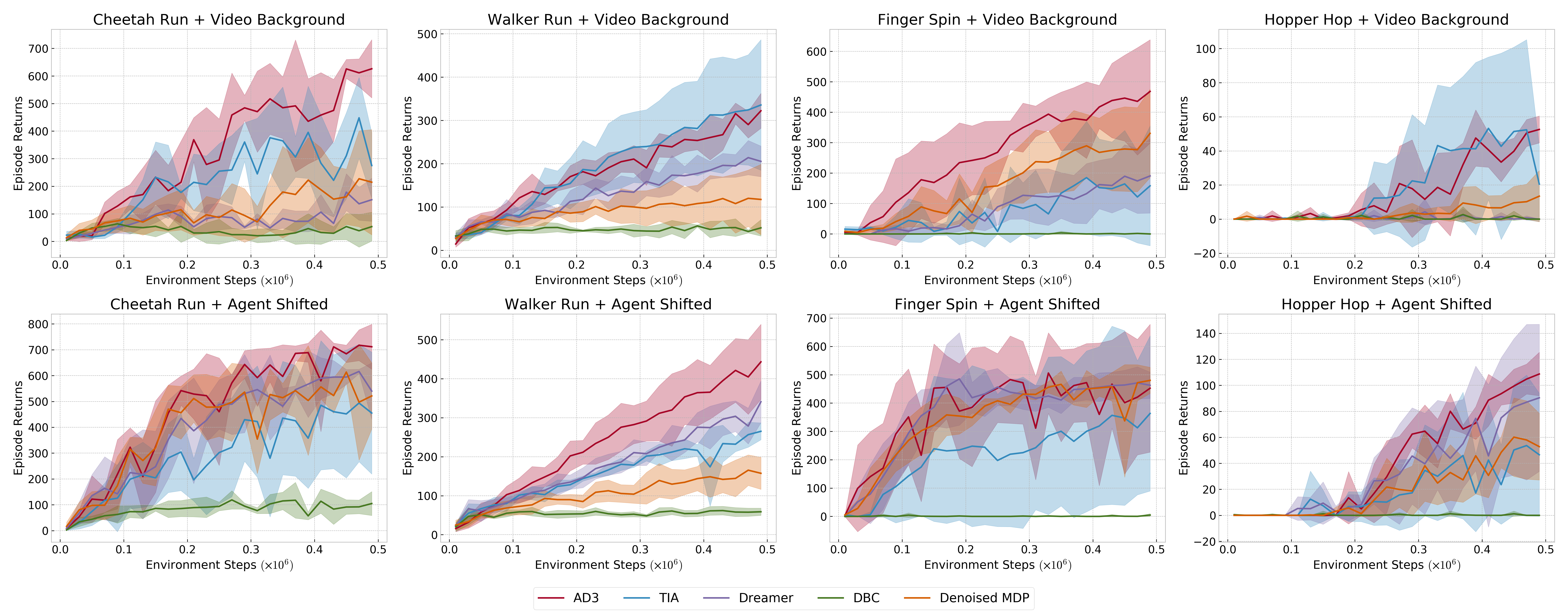}}
\vskip -0.1in
\caption{Performance evaluation of AD3 and baselines over 4 seeds across four visual control tasks, each equipped with two representative distractors: Agent Shifted and Natural Video Backgrounds. The solid curves and the shaded region indicate the average episodic returns and the standard error across different runs, respectively. AD3 is the only method
that consistently performs well across all tasks and distractor variants.}
\label{fig: performance}
\end{center}
\vskip -0.15in
\end{figure*}

\section{Experiments}
We conduct experiments to answer the following scientific research questions:
\begin{enumerate} 
    \item \textbf{RQ1}: How well does AD3 perform in environments with visual inputs that contain complex distractors? 
    \item \textbf{RQ2}: How important are the implicit actions for filtering out task-irrelevant information in visual RL tasks? 
    \item \textbf{RQ3}: What impact do different design choices in AD3 and IAG have on the experimental results?
    \item \textbf{RQ4}: Do implicit actions learned by the IAG module possess interpretable semantics? 
\end{enumerate}

\subsection{Environments and Tasks}
\label{sec: Environments and Tasks}

We evaluate AD3 and the baselines on four tasks from the DeepMind Control Suite \cite{tassa2018deepmind}: \textit{Cheetah Run}, \textit{Walker Run}, \textit{Finger Spin}, and \textit{Hopper Hop}. As mentioned in \cref{sec: Introduction}, distractors existing in visual control tasks can be divided into two main categories: heterogeneous distractors, revealing totally distinct visual semantics from the behavior of the agent, and homogeneous ones, which have similar morphology with the real agent that can be indistinguishable for the policy on which one to control. We select each of a representative for these two main categories, respectively: Natural Video Backgrounds (NBV) based on the experimental design of DBC \cite{DBLP:conf/iclr/0001MCGL21} where we replace the background of the observations with "driving car" videos in Kinetics dataset \cite{kay2017kinetics}, and Agent Shifted (AS) based on the InfoPower work \cite{DBLP:conf/iclr/BharadhwajBEL22}, where the background contains pre-recorded motion of a morphologically similar agent that is being controlled. The details of the environment design are in \cref{appendix: Environments and Tasks}.

We compare AD3 with several Visual Reinforcement Learning methods. For model-free methods, we select \textbf{DBC} \cite{DBLP:conf/iclr/0001MCGL21}, which learns task-relevant state representation with bisimulation metric. For model-based methods, we select \textbf{Dreamer} \cite{hafner2019dream}, which learns a world model and optimizing policy in the latent space with imagination, \textbf{TIA} \cite{DBLP:conf/icml/FuYAJ21}, which models both task-relevant and -irrelevant dynamics to enhance policy performance in distraction environments, and \textbf{Denoised MDP} \cite{DBLP:conf/icml/0001D0IZT22}, which factorizes the latent state based on controllability as well as reward-relevance and learns the dynamic model for each factor.

\subsection{How well does AD3 
 perform in environments with visual inputs that contain complex distractors? }
 \label{section_experiment_performance}

 In \cref{fig: performance}, we show the performance evaluation results of AD3 and baselines. AD3 is the only method that consistently performs well across all tasks and distractor variants, showcasing its robustness in handling different types of visual distractors. TIA exhibits significant variance across almost all tasks and distractors. This variance can be attributed to the unstable learning process of the injected additional loss objectives, and the heuristic assumptions behind them may become incorrect under AS setting. A more detailed analysis of TIA's weakness when dealing with homogeneous distractors can be found in \cref{appendix: Weakness of TIA}. Denoised MDP factorizes the observation into different states, but the visualization results of factorization are not as expected on most tasks, causing the learning failure. 

While struggling in the context of NBV, it is surprising that Dreamer outperforms other methods, except AD3, in tasks containing AS distractors. Since Dreamer encodes all the image information into the learned latent state, we suppose that, although task-irrelevant parts are incorporated, at least Dreamer retains all the usable information. This will ensure the lower bound of the performance, particularly in scenarios with no complex heterogeneous distractors like natural videos under AS setting.

\subsection{How important are the implicit actions for filtering out task-irrelevant information in visual RL tasks? }

\label{section: action_type}
To study the impact of the implicit action learned by IAG for distinguishing task-relevant and irrelevant components, we conduct the following experiments on two tasks with Agent Shifted distractors. We replace the implicit action $\hat{a}^-$ in AD3 with other action choices to serve as representing the semantics of the distractor's dynamics, and construct the separated world models. The model learning process is completely the same as AD3 by formulating \cref{AD3ELBO} with the chosen distractor action. No extra loss is added in, so the performance will purely reflect the influence of different choices of distractor action itself.

\begin{figure*}[t]
\begin{center}
\centerline{\includegraphics[width=\textwidth]{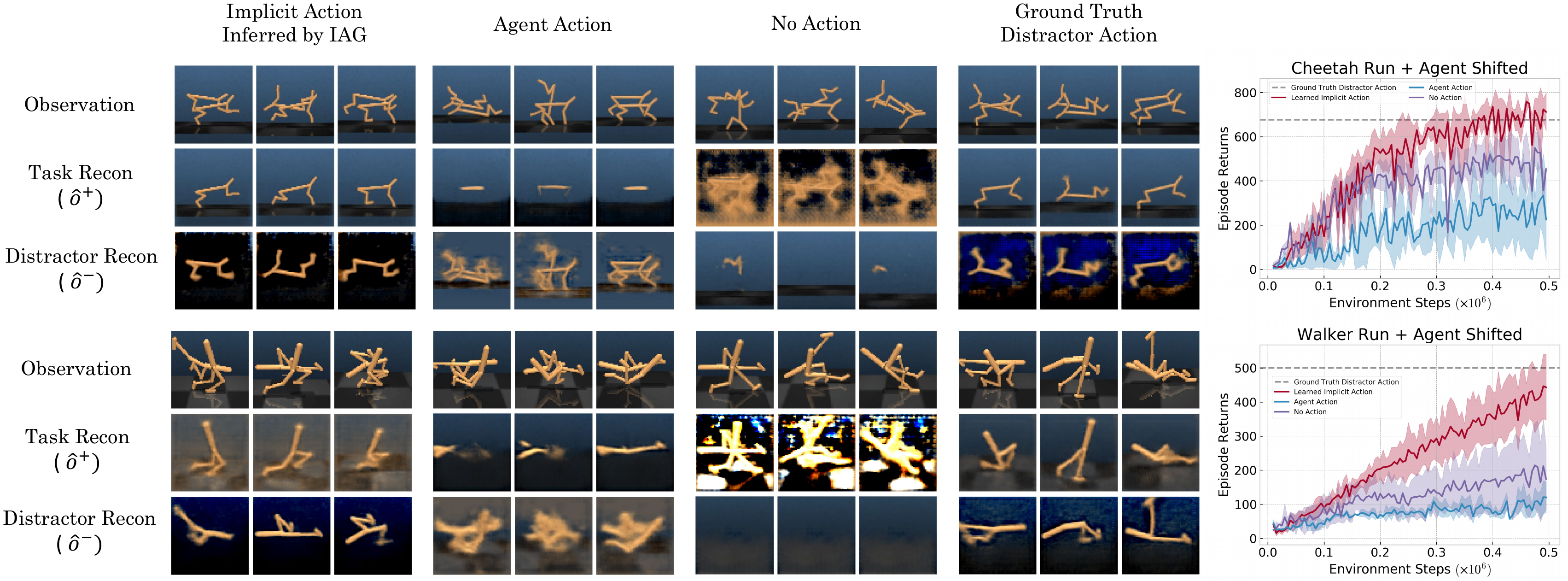}}
\vskip -0.05in
\caption{Performance and reconstruction results for different semantics of the observation, when using 4 distinct types of distractor actions for learning the task-irrelevant model under the Agent Shifted setting. Each experiment involves two tasks: \textit{Cheetah Run} and \textit{Walker Run}. When employing the ground truth action of the distractor, effective separation between the primary agent and the shifted distractor is achieved, and so do implicit actions learned by IAG, underscoring the efficacy of the implicit actions and their semantic consistency with actual distractor actions. Using agent action leads to a reversal in the representation of the two components, and the reconstructed $\hat{o}^+$ contains little task-related information. The "no action" approach tends to preserve most of the information in the task-relevant part, causing failure in the objective of distractor filtering.}
\label{fig: action_type}
\end{center}
\vskip -0.2in
\end{figure*}

The optional action to be used as the variant of $\hat{a}^-$ includes:
\begin{itemize}
     \item \textbf{Agent Action}, which is equivalent to the  practice in TIA  \cite{DBLP:conf/icml/FuYAJ21} according to TiMDP assumption. 
     \item \textbf{No Action}, which is employed in EX-BMDP  \cite{efroni2021provable}, IsoDream  \cite{pan2022iso} and SeMAIL  \cite{wan2023semail}, assuming that the transition of distractor is not conditioned on any action.
     \item \textbf{Ground Truth Distractor Action $\hat{a}^{-*}$}. Under the AS setting, we can access the actions actually executed by the shifted agent (from the loaded action sequences), which serve as ground truth for the distractor's motion. The performance of $\hat{a}^{-*}$ represents the upper bound for AD3. We regard it as an oracle for comparison.
\end{itemize}

By separated model learning conditioned on the same $a^+$ and different ${a}^-$, we can obtain different task-relevant and irrelevant parts $z^+$ and $z^-$. We visualize the observation reconstruction results $\hat{o}^+$, $\hat{o}^-$ from  $z^+$, $z^-$ as well as the evaluation performance in \cref{fig: action_type}. 

The ground truth distractor action exhibits exceptional test return and disentanglement performance. This result underscores the notion that near-optimal task / background separation can be achieved as long as we strive for the precise estimation of $\hat{a}^{-}$ to the greatest extent possible. Furthermore, implicit actions of distractors inferred by IAG also show fantastic performance on separating task-relevant main agent and the irrelevant shifted one, and scores even higher test return compared to the ground truth action on \textit{Cheetah Run} task. This demonstrates that the implicit actions learned by IAG exhibit semantic consistency with the ground truth actions of the homogeneous distractor.

The other two variants show poor ability in terms of both performance and separation. Employing agent action in the task-irrelevant model can lead to a collapse in disentanglement, resulting in a complete reversal of the two components. This is attributed to the task-irrelevant part learning task-related behavior when using agent action to model it. Moreover, the "No Action" approach consistently preserves all the information within the task-relevant state and fails to filter out the distractors. It is so hard to capture the semantics of background distractors when relying solely on states with no assistance of actions, especially for homogeneous ones. Previous methods then resort to introducing extra loss to enforce the task-irrelevant part to incorporate more information from the image \cite{wan2023semail}, which is unstable and unsuitable for AS setting. In contrast, implicit actions of distractors learned by IAG make our method much superior. Such a huge difference between the above methods illustrates the significant role of the implicit action learned by IAG, in terms of distractor filtering.

\subsection{What impact do different design choices in AD3 and IAG have on the experimental results?}
\label{sec: ablation study}
We focus on validating whether the critical designs in IAG benefit the performance and task-irrelevance distinguishing ability, including quantization technique, usage of agent action in TAID, and loss objectives. We remove each of these designs and run experiments on four different tasks, and the results are shown in \cref{tb: ablation}. Note that "No Categorical Variables" indicates using VQ.

\begin{table*}[t]
\vskip -0.15in
\begin{centering}
\caption{Ablation study on critical designs in IAG. We present the mean and std of final performance on 10 test trajectories over 4 seeds.}
\vskip 0.1in
\label{tb: ablation}
\begin{small}
\begin{sc}
\begin{tabular}{lcccc}
\toprule
Method & Cheetah Run + NBV & Walker Run + AS & Finger Spin + NBV & Hopper Hop + AS\\
\midrule
No Categorical Variables         & 276 $\pm$ 176   & 245 $\pm$ 159  &  381 $\pm$ 81  &  3 $\pm$ 4 \\
No Agent Action          & 559 $\pm$ 135 & 240 $\pm$ 122  &  418 $\pm$ 67  & 59 $\pm$ 45 \\
No Cycle Consistency    & 370 $\pm$ 217 & 240 $\pm$ 141  &  408 $\pm$ 76  & 62 $\pm$ 66 \\
No Difference Recon  & 442 $\pm$ 190 & 293 $\pm$ 154   &  \textbf{549 $\pm$ 273}  & 36 $\pm$ 51 \\
No One-step Recon  & \textbf{624 $\pm$ 164} & \textbf{422 $\pm$ 86}   &  273 $\pm$ 193  & 16 $\pm$ 23 \\
AD3    & \textbf{658 $\pm$ 104} & \textbf{435 $\pm$ 85}  &  491 $\pm$ 189 & \textbf{99 $\pm$ 31} \\
\bottomrule
\end{tabular}
\end{sc}
\end{small}
\end{centering}
\vskip -0.02in
\end{table*}

Evidently, removing any of these design components results in a performance drop, except in \textit{Finger Spin} + NBV, where the removal of difference reconstruction surprisingly increases the performance. Since most ablation methods perform relatively well on this task, we hypothesize that \textit{Finger Spin} + NBV is relatively easy and does not require pixel-level precise reconstruction. Moreover, we can observe the intuitive result that cycle consistency seems to be the most important one within the three loss objectives. The removal of agent action in TAID will trigger a huge performance drop, 
indicating the significance of agent actions for bottlenecking the implicit actions of distractors in TAID. The employment of categorical variables exhibits the greatest importance. We conclude that all of the design components in IAG regarding logical framework (usage of agent action), loss objectives, and representation form, have been found to be crucial.

\subsection{Do implicit actions learned by the IAG module possess interpretable semantics? }

Using FIAD within the IAG framework, we can generate dynamics prediction results. Specifically, we can use IAG to infer implicit action $\hat{a}_t^{-}$ by TAID for $T$ steps from $t=0$, or directly sample a series of implicit actions. The obtained $\hat{a}^{-}_{0: T}$, along with the original $a^{+}_{0: T}$, can be used to generate forward predictions from $s_{0}$ using FIAD. The results are then decoded to reconstruct the predicted observations $\hat{o}_{1:T}$ at each step. This enables us to visualize and interpret the impact of the learned implicit actions of distractors.

\begin{figure}[h]
\begin{center}
\includegraphics[width=\columnwidth]{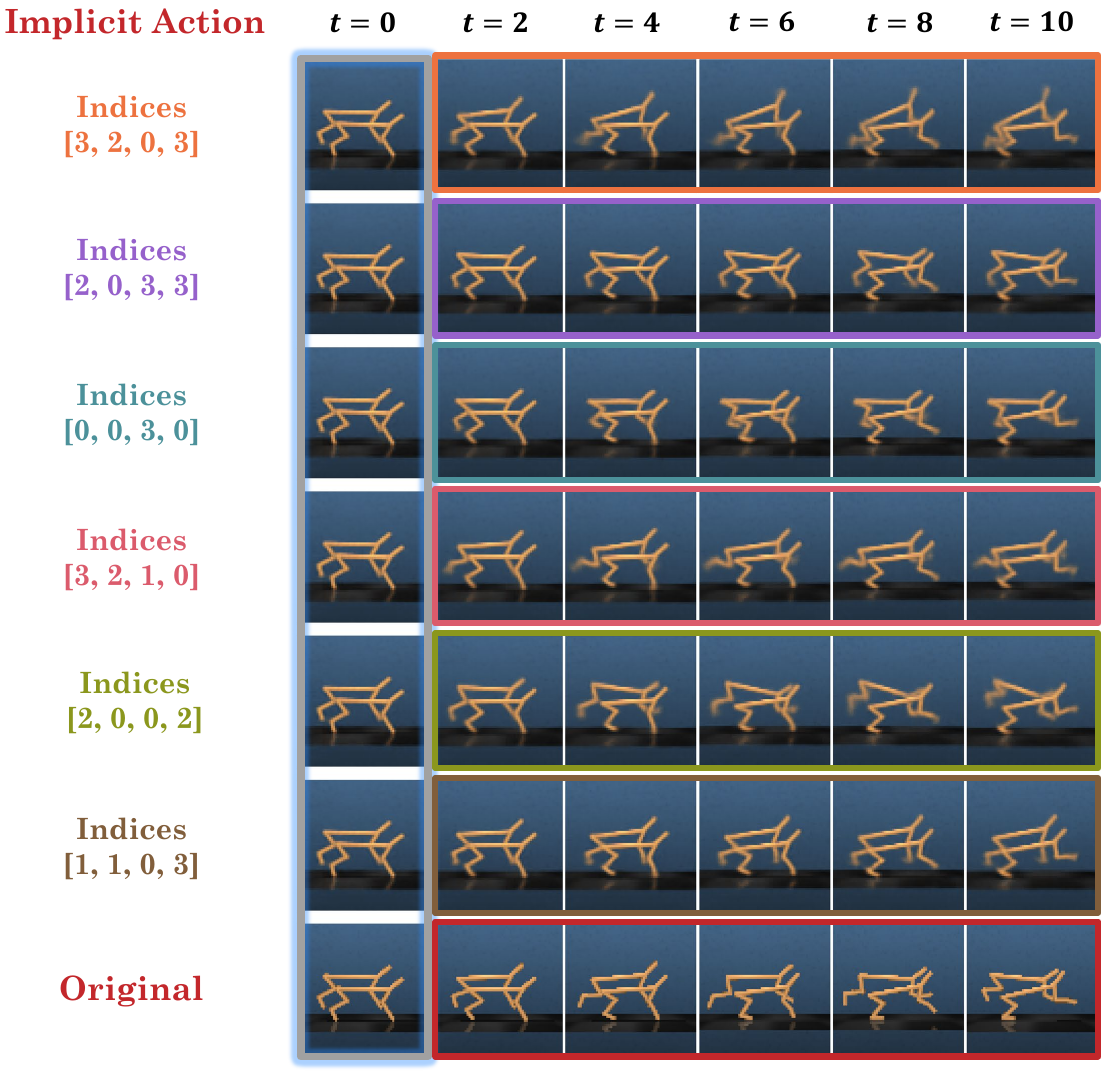}
\vskip -0.03in
\caption{Effects of different implicit actions in \textit{Cheetah Run} + AS (the size of implicit actions is 4). Conditioned on the same initial observation and identical agent action sequences from the original trajectory, we rollout FIAD for 10 steps using 6 different implicit actions of the distractor sampled from the categorical action space. These implicit actions, each represented by 4 one-hot codes with indices indicating active positions in the categorical variables, generate 6 distinct synthetic trajectories where the shifted agent exhibits different behaviors. This demonstrates that the learned implicit action space is rich in the semantic information of the underlying distractors.}
\label{fig: action_semantics}
\end{center}
\vskip -0.2in
\end{figure}

\textbf{Semantics of Different Implicit Actions} To gain insight into the semantics of the learned actions, we sample from the learned space of categorical variables to generate several distinct implicit actions of the distractor. Using FIAD, each sampled action is executed consecutively for $T$ steps to produce forward transition results in the context of AS distractors. The visualization results are shown in \cref{fig: action_semantics}. We observe that different implicit actions may exhibit distinct transition semantics, demonstrating that the implicit actions learned by IAG can effectively represent various behaviors of the distractors. This helps explain why AD3 excels at capturing the dynamics of the underlying distractors. 

\begin{figure*}[tbp]
\begin{center}
\centerline{\includegraphics[width=0.72\textwidth]{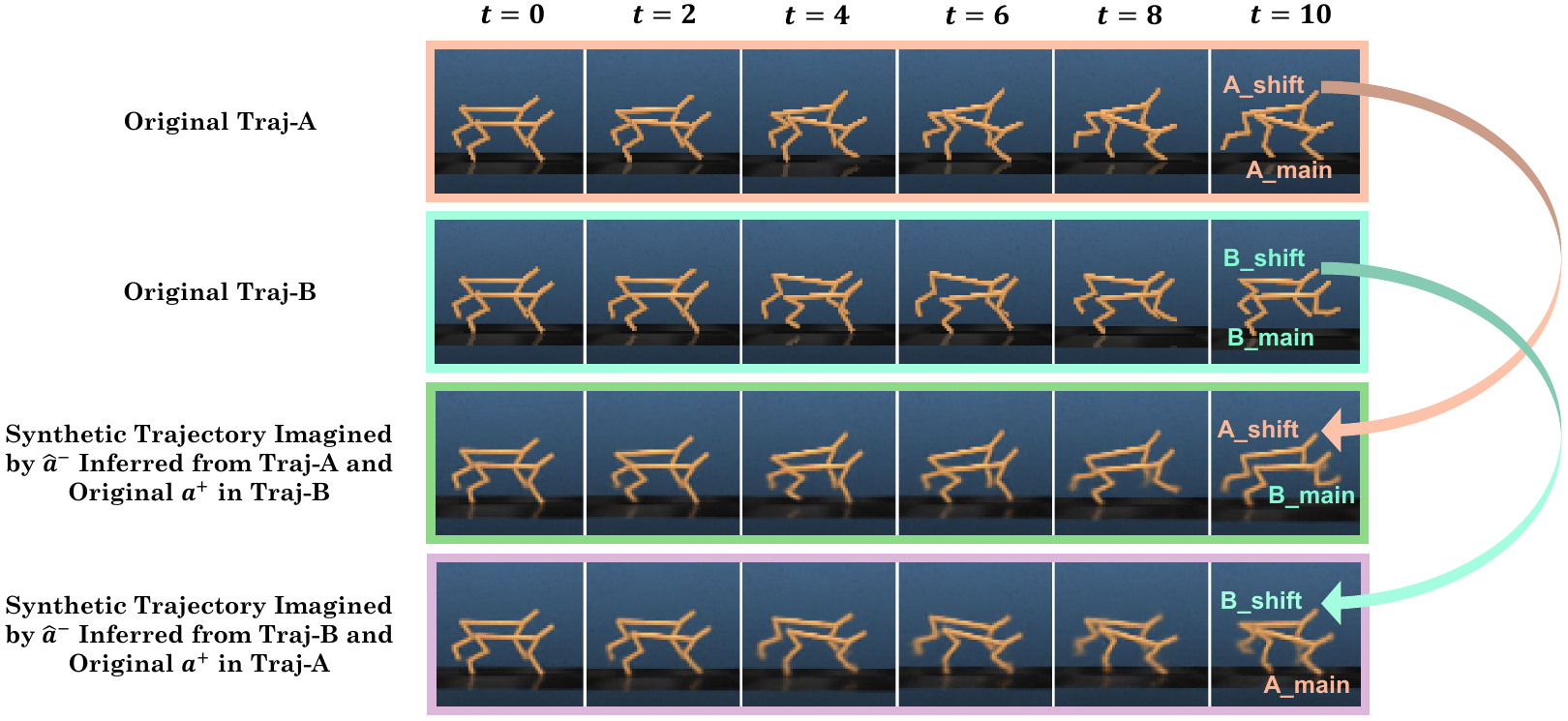}}
\caption{Forward imagination on one trajectory using implicit actions inferred from another. Using FIAD in IAG, we generate a synthetic trajectory with agent actions from Traj-B and implicit distractor actions inferred from Traj-A by TAID. The imagined trajectory exhibits the behavior of the shifted agent in Traj-A and the controllable agent in Traj-B, without incorporating task-relevant semantics from Traj-A into Traj-B. A similar result is observed when we reverse the two trajectories. This illustrates the effective disentanglement of learned implicit actions from original agent actions.}
\label{fig: imagine_cross_trajectory}
\end{center}
\vskip -0.2in
\end{figure*}

\textbf{Implicit Actions are Disentangled from Agent Actions}
An ideal implicit distractor action $\hat{a}^-$ should not only contain the transition information about the distractor but also possess no task-relevant signal, e.g., the semantics of the controllable agent under the AS setting. We verify such disentanglement by conducting forward dynamics prediction on one trajectory using implicit distractor actions inferred from another, which is visualized in \cref{fig: imagine_cross_trajectory}. Evidently, the final result displays a composite behavior of different agents originating from distinct trajectories, indicating effective removal of task-relevant semantics within the implicit distractor actions. Such empirical results validate the interpretation of TAID in \cref{eq:IAG-model-interpretation}, where the inferred implicit action $\hat{a}^-$ is completely disentangled from agent action $a^+$. The remarkable performance of separated world model learning can thus be achieved.

\section{Discussion}
Previous methods for eliminating distractors in visual control primarily focus on heterogeneous distractors, leaving homogeneous ones largely unexplored. To address this issue, we introduce Implicit Action Generator (IAG) to infer implicit actions associated with visual distractors, and present a novel algorithm, \emph{implicit \textbf{A}ction-informed \textbf{D}iverse visual \textbf{D}istractors \textbf{D}istinguisher} (AD3), leveraging the implicit actions to facilitate the training of separated world models.  Our method demonstrates superior performance across various visual control tasks featuring both heterogeneous and homogeneous distractors. Empirical evidence also underscores the indispensable role played by implicit actions learned through IAG and their interpretable semantics.

AD3 particularly excels in handling homogeneous distractors such as shifted agents. This aspect has rarely been explored before but can more accurately reflect the policy's ability to extract task-relevant information than heterogeneous distractors, as homogeneous ones lack easily distinguishable visual features and thus present a greater challenge. Our method achieves near-perfect separation of homogeneous distractors, a task that is challenging for both visual methods like SAM \cite{kirillov2023segment} and previous RL methods. We provide visualization results of task-relevant and irrelevant reconstructions in \cref{sec: Additional Visualization Result}, and more extensive dynamic results (videos) can be found at \href{https://sites.google.com/view/ad3-iag}{https://sites.google.com/view/ad3-iag}.

For distractor-eliminating problems in visual RL, we are the first to introduce the inference of implicit actions from existing distractors and utilize them to construct separated world models. This methodology captures the inherent nature of the \textbf{control} problem itself, rather than relying on additional visual elements or heuristic human bias.

Furthermore, the notion of inferring implicit actions from videos holds broad applicability. The recently proposed Genie \cite{bruce2024genie} also infers latent actions from image sequences (videos). However, our approach distinguishes itself in that we infer implicit actions of dynamic transitions within the environment, rather than inferring the actions of the agent itself, as done by Genie and VPT \cite{baker2022video}. This broadens the scope of action inference in RL, with potential implications for areas such as video prediction and world model learning. For instance, the semantics of numerous actionless in-the-wild videos could potentially be unified by the underlying implicit actions inferred. We leave these explorations for future work.

\section*{Acknowledgements}

We would like to thank Shaowei Zhang, Bowen Zheng, Minghao Shao, Kaichen Huang, Haihang Sun, and Ruying Chen for valuable discussions. This work was supported by National Science and Technology Major Project (2022ZD0114805).

\section*{Impact Statement}

This paper aims to advance the field of machine learning by addressing challenges in visual reinforcement learning, particularly the failure to learn expected behaviors in the presence of irrelevant distractors in visual observations, such as those encountered in self-driving and motor control. Our proposed method mitigates these issues, potentially yielding significant positive impacts on industry and security.


\bibliography{AD3_paper}
\bibliographystyle{icml2024}

\newpage
\appendix
\onecolumn
\section{Derivations}
\label{appendix: deriv}

Prior research in model-based reinforcement learning \cite{hafner2019learning, hafner2019dream} has employed the variational evidence lower bound (ELBO) or, more generally, the information bottleneck objective \cite{tishby2000information, alemi2016deep}, to encourage model states to predict observations and rewards, while limiting the volume of information contained within these states. Based on the Implicit-Action Block MDP assumption, we integrate the implicit action $a^-$ into the conditioned-on term of the objective function in  \cite{hafner2019learning, hafner2019dream}, resulting in a new objective:
\begin{equation}
\label{original MI objective}
    \max \text{I}(z_{1:T}, (o_{1:T}, r_{1:T})|a_{1:T}, a^{-}_{1:T}) - \beta \text{I}(z_{1:T}, i_{1:T}|a_{1:T}, a^{-}_{1:T})
\end{equation}
Here, $i_t$ are dataset indices such that $p(o_t|i_t) = \delta(o_t-o_t^{\prime})$ as in  \cite{alemi2016deep}. The lower bound of the first term can be obtained using the definition of mutual information as well as the non-negativity of KL-divergence:
\begin{align}
    &\text{I}(z_{1:T}, (o_{1:T}, r_{1:T})|a_{1:T}, a^{-}_{1:T}) \\
    =~& \mathbb{E}_{p(o_{1:T}, r_{1:T}, z_{1:T}, a_{1:T}, a^{-}_{1:T})}\left[\ln p(o_{1:T}, r_{1:T}|z_{1:T}, a_{1:T}, a^{-}_{1:T})-\ln p(o_{1:T}, r_{1:T}|a_{1:T}, a^{-}_{1:T})\right]\label{eq:reduce_const}\\
    \overset{+}{=}~& \mathbb{E}_{p(o_{1:T}, r_{1:T}, z_{1:T}, a_{1:T}, a^{-}_{1:T})}\left[\ln p(o_{1:T}, r_{1:T}|z_{1:T}, a_{1:T}, a^{-}_{1:T})\right] \\
    \ge~& \mathbb{E}_{p(o_{1:T}, r_{1:T}, z_{1:T}, a_{1:T}, a^{-}_{1:T})}\left[\ln p(o_{1:T}, r_{1:T}|z_{1:T}, a_{1:T}, a^{-}_{1:T})\right] \nonumber\\
    -~& \mathbb{D}_{\text{KL}}\left(p(o_{1:T}, r_{1:T}|z_{1:T}, a_{1:T}, a^{-}_{1:T}) \| \prod_{t=1}^T q(o_t|z_t)q(r_t|z_t)\right)\\
    =~& \mathbb{E}_{p(o_{1:T}, r_{1:T}, z_{1:T}, a_{1:T}, a^{-}_{1:T})}\left[\sum_{t=1}^T\ln q(o_t|z_t)q(r_t|z_t)\right]\\
    =~& \mathbb{E}_{p(o_{1:T}, r_{1:T}, a_{1:T})} \left[ \sum\limits_{t=1}^{T} \mathbb{E}_{p(z_{t}^+|o_{1:t},a_{1:t-1})p(z_{t}^-|o_{1:t}, a^-_{1:t-1})} \left[\ln q(o_{t}|z_{t}^+,z_{t}^-) + \ln q(r_t|z_{t}^+)\right]\right]\label{eq:recon}
\end{align}
We drop the second term in the equality (\ref{eq:reduce_const}) since it can be regarded as constant for the representation model. The equality (\ref{eq:recon}) is obtained from the decoupling of the latent state and the decomposition of dynamics as assumed in IABMDP.

We then upper bound the second term by using the non-negativity of the KL-divergence and the IABMDP assumption:
\begin{align}
&\text{I}(z_{1:T},i_{1:T}|a_{1:T}, a^-_{1:T})\\
=~ & \mathbb{E}_{p(o_{1:T}, r_{1:T}, z_{1:T}, i_{1:T}, a_{1:T}, a^-_{1:T})}\left[\sum_{t=1}^T\ln p(z_t|z_{t-1},a_{t-1},a^{-}_{t-1}, i_t)-\ln p(z_t|z_{t-1}, a_{t-1}, a^{-}_{t-1})\right]\\
=~ & \mathbb{E}_{p(o_{1:T}, z_{1:T}, a_{1:T}, a^-_{1:T})} \left[\sum_{t=1}^T\ln p(z_t|z_{t-1},a_{t-1},a^{-}_{t-1}, o_t)-\ln p(z_t|z_{t-1}, a_{t-1}, a^{-}_{t-1})\right]\\
\leq ~& \mathbb{E}_{p(o_{1:T}, z_{1:T}, a_{1:T}, a^-_{1:T})} \left[\sum_{t=1}^T\ln p(z_t|z_{t-1},a_{t-1},a^{-}_{t-1}, o_t)-\ln q(z_t|z_{t-1}, a_{t-1}, a^{-}_{t-1})\right]\\
=~&\mathbb{E}_{p(o_{1:T}, z_{1:T}, a_{1:T}, a^-_{1:T})}
\left[\sum\limits_{t=1}^{T} \ln\frac{p(z_{t}^+|o_{t},z_{t-1}^+,a_{t-1})p(z_{t}^-|o_{t},z_{t-1}^-,a^-_{t-1})}{q(z_{t}^+|z_{t-1}^+,a_{t-1})q(z_{t}^-|z_{t-1}^-, a^-_{t-1})}\right]\\
=~&\mathbb{E}_{p(o_{1:T},a_{1:T}, a^-_{1:T})}\Bigg[\mathbb{E}_{p(z_{1:T}^+|o_{1:T},a_{1:T})}\left[\sum\limits_{t=1}^{T} \ln\frac{p(z_{t}^+|o_{t},z_{t-1}^+,a_{t-1})}{q(z_{t}^+|z_{t-1}^+,a_{t-1})}\right] \nonumber \\ 
&+\mathbb{E}_{p(z_{1:T}^-|o_{1:T},a^-_{1:T})}\left[\sum\limits_{t=1}^{T} \ln\frac{p(z_{t}^-|o_{t},z_{t-1}^-,a^-_{t-1})}{q(z_{t}^-|z_{t-1}^-,a^-_{t-1})}\right]\Bigg]\\
=~&\mathbb{E}_{p(o_{1:T},a_{1:T}, a^-_{1:T})} \Bigg[\sum\limits_{t=1}^{T}\Big(\mathbb{E}_{p(z_{t-1}^+|o_{1:t-1},a_{1:t-2})}\left[\mathbb{D}_{\text{KL}}\left({p(z_{t}^+|o_{t},z_{t-1}^+,a_{t-1})}\|{q(z_{t}^+|z_{t-1}^+,a_{t-1})}\right)\right] \nonumber \\
&+\mathbb{E}_{p(z_{t-1}^-|o_{1:t-1},a^-_{1:t-2})}\left[\mathbb{D}_{\text{KL}}\left({p(z_{t}^-|o_{t},z_{t-1}^-,a^-_{t-1})}\|{q(z_{t}^-|z_{t-1}^-,a^-_{t-1})}\right)\right]\Big)\Bigg]
\end{align}
The original objective \ref{original MI objective} can thus be lower bounded. In practice, obtaining the ground truth $a^-$ is infeasible. Thus, we substitute it with the implicit distractor actions $\hat{a}^-$ inferred by IAG. We use two pairs of observation decoders $q_{\phi^+}$ and $q_{\phi^-}$, forward dynamics models $q_{\theta^+}$ and $q_{\theta^-}$, and variational encoders $p_{\psi^+}$ and $ p_{\psi^-}$ for the task-relevant and irrelevant branches respectively, as well as a reward decoder $q_{\omega}$. The final objective to be optimized for separated model learning is as follows:
\begin{equation}
\label{equation_complete_world_model_learning}
\begin{split}
    \max_{\theta^{+},\theta^{-}, \psi^{+}, \psi^{-}\atop \phi^{+}, \phi^{-}, \omega} &\quad\mathbb{E}_{(o_{\tau},a_{\tau},\hat{a}^{-}_{\tau})\sim\mathcal{B}_{\pi}}\\
    &\Bigg[\sum_{t=1}^T\Big(
    \mathbb{E}_{p_{\psi^{+}}(z_{t-1}^+|o_{1:t-1},a_{1:t-2})}\left[-\mathbb{D}_{\text{KL}}\left(p_{\psi^+}(z^+_t|o_t, z^+_{t-1}, a_{t-1})||q_{\theta^+}(z^+_t|z^+_{t-1}, a_{t-1})\right)\right]\\
    &+\mathbb{E}_{p_{\psi^{-}}(z_{t-1}^-|o_{1:t-1},\hat{a}^-_{1:t-2})}\left[-\mathbb{D}_{\text{KL}}\left(p_{\psi^-}(z^-_t|o_t, z^-_{t-1},\hat{a}^-_{t-1})||q_{\theta^-}(z^-_t|z^-_{t-1},\hat{a}^-_{t-1})\right)\right]\\
    &+\mathbb{E}_{p_{\psi^{+}}(z_{t}^+|o_{1:t},a_{1:t-1})p_{\psi^{-}}(z_{t}^-|o_{1:t},\hat{a}^-_{1:t-1})}\left[\ln q_{\phi^{+}, \phi^{-}}(o_{t}|z^+_{t}, z^-_{t}) + \ln q_{\omega}(r_t|z^+_{t}) \right]\Big)\Bigg]
\end{split}
\end{equation}

\section{Implementation Details}

\subsection{Environments and Tasks}
\label{appendix: Environments and Tasks}
The DeepMind Control Suite (DMC) \cite{tassa2018deepmind} is a collection of simulated continuous control tasks covering a diverse range of environments including robotics, locomotion, manipulation, and navigation. 
We select four tasks from the DeepMind Control Suite: \textit{Cheetah Run}, \textit{Walker Run}, \textit{Finger Spin}, and \textit{Hopper Hop}.
 
\begin{figure*}[tbp]
\begin{center}
\centerline{\includegraphics[width=0.9\textwidth]{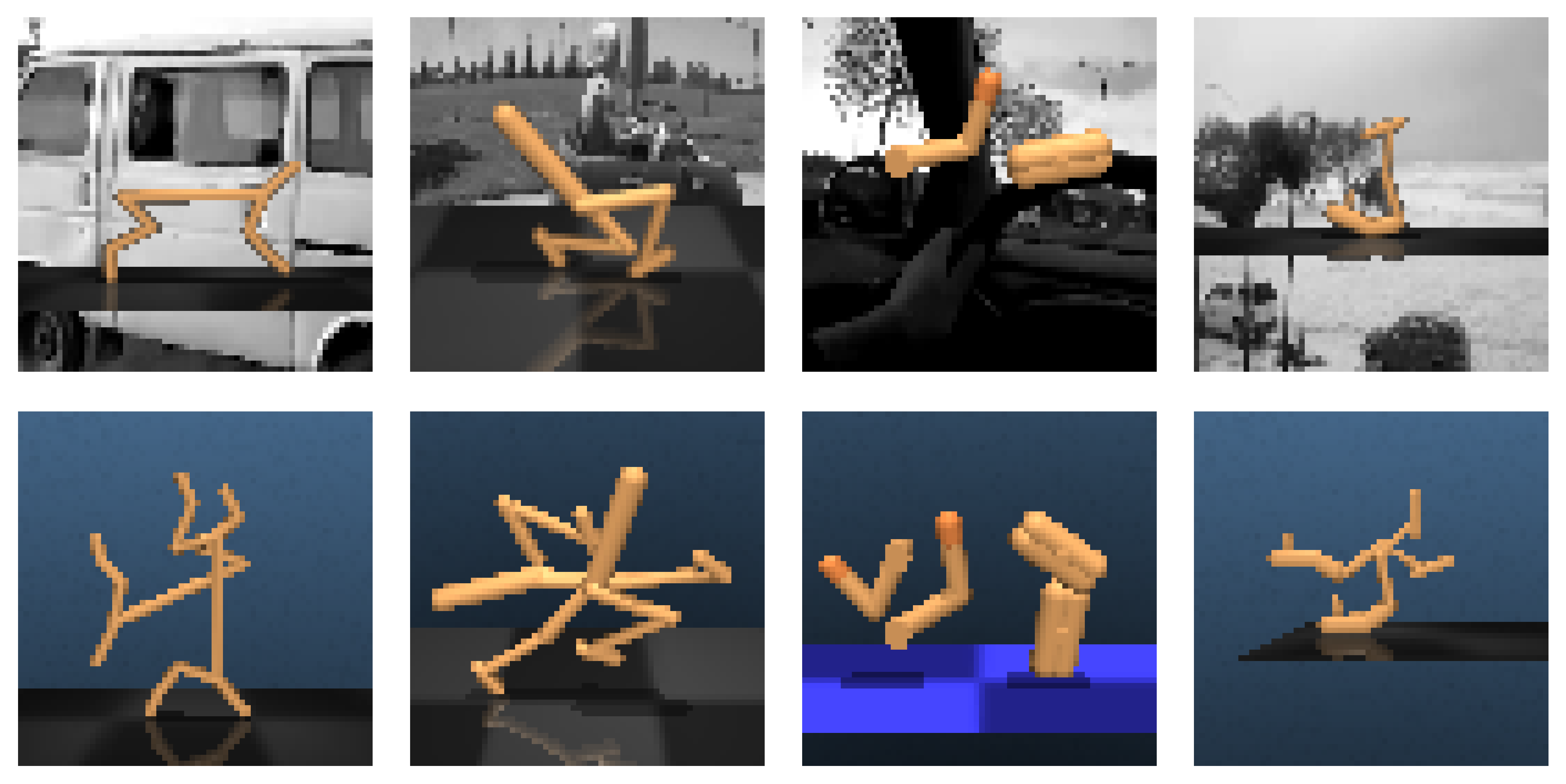}}
\vskip -0.1in
\caption{The example observations of the environments we evaluate on, featuring \textit{Cheetah Run}, \textit{Walker Run}, \textit{Finger Spin}, and \textit{Hopper Hop} from left to right. The first row displays these environments with NBV distractors, while the second row shows them with AS distractors. }
\label{fig: original_8_envs}
\end{center}
\vskip -0.3in
\end{figure*}

For each task, we introduce two types of visual distractors: Natural Video Backgrounds (NBV) and Agent Shift (AS), which are representatives of heterogeneous and homogeneous distractors, respectively. For the NBV distractor, we replace the background of the visual observations with "driving car" videos in the Kinetics dataset \cite{kay2017kinetics}. The video backgrounds are grayscale as in \cite{DBLP:conf/iclr/0001MCGL21, DBLP:conf/icml/FuYAJ21, wan2023semail}. With respect to the AS distractor, we integrate the motion of a
morphologically similar agent to the one that is being controlled to the observation background based on the InfoPower work \cite{DBLP:conf/iclr/BharadhwajBEL22}, where the motion of this distractor agent is loaded from a training buffer of TIA on the same task. The policy cannot access the actions actually executed by the background agent, except for the variant method "Ground Truth Distractor Action $\hat{a}^{-*}$" introduced in \cref{section: action_type}, where we use $\hat{a}^{-*}$ as an oracle for comparison. The example observations of these environments we evaluate are shown in \cref{fig: original_8_envs}.

\subsection{Implicit Action Space Using Categorical Variables}
\label{appendix: one-hot coding implementation}

To prevent shortcuts during the observations reconstruction process, we need to bottleneck the information flow from the combination of adjacent frames $o_t,o_{t+1}$ and agent action $a_t$ to the predicted next frame $\hat{o}_t$. While previous works have employed vector quantization techniques for this purpose, we introduce an extremely sparse implementation termed categorical implicit action, which demonstrates empirical effectiveness in our experiments. We follow the practice of DreamerV2 \citep{DBLP:conf/iclr/HafnerL0B21} to implement the categorical variables technique, and each implicit action is represented by a vector of multiple categorical (one-hot) variables.

Consider implicit actions with size $d$. We implement the last layer of the implicit action generator (IAG) module as a dense layer with dimension $d^2$, which outputs the logits of shape $N\times d^2$, where $N$ is the batch size. Then we reshape the logits as $N\times d\times d$. Subsequently, for one sample $z$ (shape: $d\times d$) in the batch, a softmax activation function is applied along the last dimension of the logits to output $d$ probability vectors $p_{\hat{a}^-}^i$, each with size $d$. We build a categorical distribution based on each probability vector $p_{\hat{a}^-}^i$ and sample a one-hot vector from each distribution. We concatenate all $d$ one-hot vectors to form the final implicit action $\hat{a}^-$. 

\begin{equation}\label{eq:implicit_categorical_action}
	\begin{aligned}
	p_{\hat{a}^-}^i &= \text{Softmax}(z_i), \quad \text{where } i=1, 2,\cdots, d\\
	\hat{a}^- &= \Big[\text{Sample}(\text{Categorical}(p_{\hat{a}^-}^1)); \text{Sample}(\text{Categorical}(p_{\hat{a}^-}^2)); \cdots; \text{Sample}(\text{Categorical}(p_{\hat{a}^-}^d))\Big]
	\end{aligned}
\end{equation}

Since each implicit action is a vector of $d$ categorical variables and each categorical variable is in size $d$, we can represent an implicit action by using $d$ indices which respectively indicates the active position in each categorical variable, as is the case in \cref{fig: action_semantics} where implicit action size $d=4$. Furthermore, we follow the practice of straight-through gradients \cite{bengio2013estimating} mentioned in DreamerV2 to implement the optimization process of categorical variables.

\subsection{Pseudo Code}
The pseudo-code of our proposed AD3 is provided in \cref{alg:AD3}.
\begin{algorithm}
   \caption{Training Procedure of AD3}
   \label{alg:AD3}
\begin{algorithmic}
    \STATE {Initialize replay buffer $\mathcal{B}_{\pi}$, IAG, forward dynamics model $q_{\theta^+}, q_{\theta^-}$, variational encoder $p_{\psi^+}, p_{\psi^-}$, observation decoder $q_{\phi^+}, q_{\phi^-}$, reward decoder $q_{\omega}$, policy $\pi$.}
    \FOR{each time step $t=1\cdots T$}
        \STATE {\textcolor{red}{// Rollout trajectories}}
        \STATE {Infer the task-relevant latent state $z_t^+\sim p_{\psi^+}(\cdot| o_t, z_{t-1}^+, a_{t-1})$}
        \STATE {Sample action from policy $a_{t}\sim \pi(\cdot| z_t^+)$}
        \STATE {Execute action and get the next observation $o_{t+1}\leftarrow \text{env.step}(a_t)$}
    \ENDFOR
    \STATE {Add samples into the replay buffer $\mathcal{B}_{\pi}\leftarrow \mathcal{B}_{\pi}\cup \{(o_t, a_t, r_t)_{t=1}^T\}$}
    \FOR{training iteration $i=1\cdots \text{It}$}
        \STATE {\textcolor{red}{// Learn implicit actions}}
        \STATE {Sample minibatch $(o_{1:T}, a_{1:T-1})_{1:b}$ from the buffer $\mathcal{B}_{\pi}$}
        \STATE {Update IAG with~\cref{eq:IAG}}
        \STATE {Infer the implicit actions $\hat{a}^-_{1:T-1}$ using IAG and store them in the buffer $\mathcal{B}_{\pi}$}
        \STATE {\textcolor{red}{// Learn separated world models}}
        \STATE {Sample minibatch $(o_{1:T}, a_{1:T-1}, \hat{a}^-_{1:T-1}, r_{1:T})_{1:b}$ from the buffer $\mathcal{B}_{\pi}$}
        \STATE {Update $q_{\theta^+}, q_{\theta^-}$, $p_{\psi^+}, p_{\psi^-}$, $q_{\phi^+}, q_{\phi^-}$, $q_{\omega}$ with \cref{equation_complete_world_model_learning}}
        \STATE {\textcolor{red}{// Optimize policy}}
        \STATE {Imagine trajectory rollouts by policy $\pi$, using the task-relevant forward dynamics model $q_{\theta^+}$ and the reward decoder $q_{\omega}$}
        \STATE {Update the policy $\pi$ to maximize the cumulative rewards.}
    \ENDFOR
\end{algorithmic}
\end{algorithm}

\subsection{Networks and Hyperparameters}
\label{appendix: implementation networks and hyperparameters}
\subsubsection{Learning IAG} 
IAG comprises TAID, which infers the implicit action $\hat{a}_t^-$ from two adjacent image frames $o_t$, $o_{t+1}$ and the agent action $a_t$, and FIAD, which predicts the next frame $\hat{o}_{t+1}$ for consistency using $o_t$, $a_t$, and the implicit action $\hat{a}_t^-$. We elaborate on the implementation details of each component and explain the training protocol of IAG.

\textbf{TAID} We first use a convolutional image encoder to encode the augmented image frames $o_t$ and $o_{t+1}$. Next, additional convolutional layers fuse the encoded features of $o_t$ and the agent action $a_t$, as well as the features of $o_{t+1}$. An MLP then embeds these fused features into a low-dimensional latent state. The resulting embeddings are considered as logits for implicit actions, and we employ techniques described in \cref{appendix: one-hot coding implementation} to sample the inferred implicit actions $\hat{a}_t^-$ as the output of TAID. All convolutional layers are implemented using residual blocks and batch normalization.

\textbf{FIAD} FIAD aims to ensure consistency between the decoded prediction and the ground truth. We construct independent decoders for each of the three proposed prediction tasks: cycle consistency of low-dimensional latent states, pixel-level difference reconstruction, and one-step image reconstruction. For cycle consistency, we employ a CNN as the forward dynamics model to process and integrate the features of $o_t$, the agent action $a_t$, and the implicit action $\hat{a}_t^-$, ultimately predicting the feature of the next step and computing its similarity with the ground-truth feature of $o_{t+1}$. We use another forward dynamics network for difference reconstruction and employ a transposed CNN to decode the predicted feature into image difference. We utilize another transposed CNN to output the reconstruction result for one-step reconstruction.

\textbf{Building Loss Objectives} For cycle consistency, we employ contrastive methods rather than cosine similarity or MSE loss between predicted and ground-truth states. We enforce the projection of the predicted image feature to distinguish between all the results in the batch and query for the corresponding ground-truth one, and minimize the contrastive loss. Moreover, we employ multi-scope difference reconstruction, which involves predicting the max-pooled versions of the original image difference at various resolutions. We follow FICC \cite{FICC23} and use binary cross-entropy loss for the two pixel-level reconstruction objectives. Furthermore, the three objectives are built and computed in a multi-step manner as in FICC \cite{FICC23} and SPR \cite{schwarzer2020data}. Specifically, we first infer the implicit actions $\hat{a}_t^-$ for each of the transition triplet $(o_t, a_t, o_{t+1})$ in a sampled short trajectory. Then, we roll out from the latent state of the first observation using the agent actions and the inferred actions to obtain predictions at each step, through the dynamics network of cycle consistency in FIAD. Finally, we build the three objectives at each step. 

\textbf{Training Details} We train IAG every 10,000 steps of environmental interactions. In each training session, we sample a batch of observations and agent actions from the collected data buffer and perform an update for IAG. Each training procedure of IAG consists of 1,000 steps of such updates. Subsequently, we infer the implicit actions of distractors for all collected trajectories using the current parameters of TAID in IAG and store them in the buffer. Thus, every time we finish training IAG, the implicit actions of the previously collected trajectories are updated. These implicit actions, along with observations and agent actions, are then used for the next round of separated model learning. 

\textbf{Hyperparameters} We use the ADAM optimizer to train IAG with batches of 64 sequences, and the sequence length for multi-step rollout to build the loss objectives is 6. The projection dimension for constrastive loss is 512. The learning rate for training IAG is 6e-4, and the activation function is ReLU. For NBV distractors, the sizes $d$ of the implicit action for \textit{Cheetah Run}, \textit{Walker Run}, \textit{Finger Spin} and \textit{Hopper Hop} are 12, 8, 10, and 10, respectively. For AS distractors, all environments use an implicit action size of 4. According to the implementation of categorical variables in \cref{appendix: one-hot coding implementation}, the real size of the implicit action used for building the task-irrelevant world model is $d^2$, achieved by flattening each of the one-hot vectors and concatenating them. Furthermore, when using VQ for ablation study in \cref{sec: ablation study}, the codebook length is 20, and the dimension of the vector representation is $d$, identical to the size used for categorical variables in each task.

\subsubsection{Learning Separated World Models and Policy} We use the recurrent state space model (RSSM) as the backbone of our world models and adopt the official implementation of TIA for building separated models and cooperative image reconstruction through masking techniques. The separated world model is constructed using two independent RSSMs, with agent actions and implicit actions inferred by IAG respectively serving as inputs. Other implementation details, including policy learning, optimization methods, the iterative process of data collection and training with the collected data, as well as most of the related hyperparameters, are kept identical to those in TIA. This ensures a fair comparison between our AD3 method and both Dreamer and TIA, thereby providing a stronger verification of the impact of the inferred implicit actions in distinguishing irrelevant distractors.

\section{Quantitative Results}
In \cref{tb: quantitative}, we show the quantitative results of the experimental performance in \cref{section_experiment_performance}. We present the mean and std of the final test performance on 10 episodes over 4 seeds. AD3 is the only method that consistently performs well across all tasks and distractor variants, showing robustness in handling different types of visual distractors. 

\begin{table}[h]
\centering
\caption{Performance on 4 visual control tasks with heterogeneous and homogeneous visual distractors }
\vskip 0.1in
\begin{small}
\begin{sc}
\begin{tabular}{lcccccccc}
    \toprule
    \multirow{2}{*}{Method} & \multicolumn{2}{c}{Cheetah Run} & \multicolumn{2}{c}{Walker Run} & \multicolumn{2}{c}{Finger Spin} & \multicolumn{2}{c}{Hopper Hop} \\
    \cmidrule(lr){2-3} \cmidrule(lr){4-5} \cmidrule(lr){6-7} \cmidrule(lr){8-9} 
    & NBV & AS & NBV & AS & NBV & AS & NBV & AS \\
    \midrule
    DBC & 45 $\pm$ 55 & 96 $\pm$ 46 & 50 $\pm$ 27 & 63 $\pm$ 13 & 1 $\pm$ 1 & 5 $\pm$ 9 & 0 $\pm$ 0 & 0 $\pm$ 0 \\
    Dreamer & 151 $\pm$ 78 & 540 $\pm$ 109 & 205 $\pm$ 35 & 330 $\pm$ 51 & 191 $\pm$ 122 & 448 $\pm$ 44  & 0 $\pm$ 0 & \textbf{92 $\pm$ 54}  \\
    TIA & 432 $\pm$ 172 & 474 $\pm$ 251 & \textbf{293 $\pm$ 129} & 264 $\pm$ 42 & 123 $\pm$ 201 & 352 $\pm$ 264 & \textbf{50 $\pm$ 48} & 51 $\pm$ 53  \\
    Denoised MDP & 215 $\pm$ 191 & 521 $\pm$ 128 & 117 $\pm$ 81 & 157 $\pm$ 41 & 331 $\pm$ 146 & \textbf{480 $\pm$ 46} & 13 $\pm$ 15 & 53 $\pm$ 26 \\
    AD3 (Ours) & \textbf{658 $\pm$ 104} & \textbf{653 $\pm$ 130} & \textbf{308 $\pm$ 37} & \textbf{435 $\pm$ 85} & \textbf{491 $\pm$ 189} & \textbf{515 $\pm$ 228} & \textbf{57 $\pm$ 20} & \textbf{99 $\pm$ 31} \\
    \bottomrule
\end{tabular}
\end{sc}
\label{tb: quantitative}
\end{small}
\end{table}

Moreover, our work primarily focuses on learning Implicit Action Generator (IAG) to extract actions of underlying distractors, thereby aiding in eliminating task-irrelevant factors. The construction and performance optimization of specific world models are not the main contributions of this paper, and we employ the simplest Dreamer to build the separated world models. It is worth noting that more advanced methods, such as categorical latent state space and KL balancing \cite{DBLP:conf/iclr/HafnerL0B21, NEURIPS2022_a766f56d}, can be readily incorporated to enhance overall performance even more. This underscores the plug-and-play nature of our IAG method, demonstrating its potential to be seamlessly integrated into any model-based visual RL method aimed at modeling task-irrelevant dynamics.

\section{Additional Experiments}
\subsection{Performance of AD3 on More Diverse Tasks }

\begin{figure*}[tbp]
\begin{center}
\centerline{\includegraphics[width=0.8\textwidth]{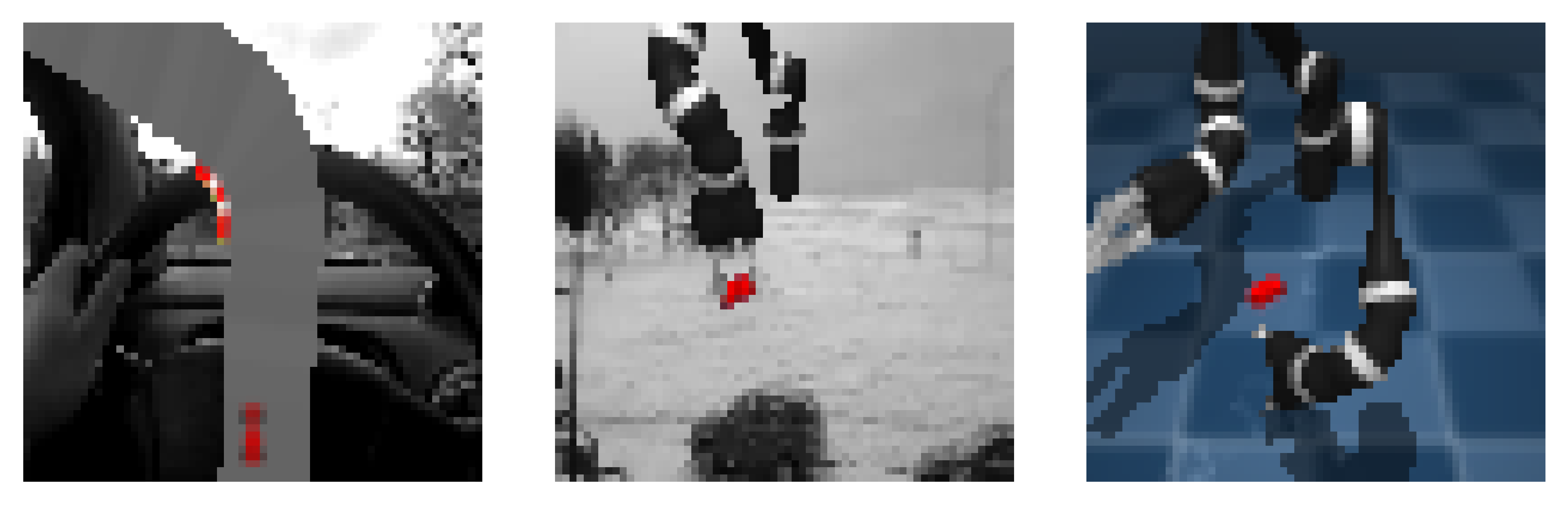}}
\vskip -0.1in
\caption{The example observations of additional evaluation environments beyond the DMC tasks with visual distractors, specifically featuring \textit{Car Racing} + NBV, \textit{Jaco Arm} + NBV, and \textit{Jaco Arm} + AS.}
\label{fig: carjaco_3_envs}
\end{center}
\vskip -0.2in
\end{figure*}

Previous work addressing pixel-based RL problems with visual distractors commonly uses DMC tasks for evaluation ( \cite{DBLP:conf/iclr/0001MCGL21, DBLP:conf/icml/FuYAJ21, DBLP:conf/iclr/BharadhwajBEL22} for NBV distractor, \cite{DBLP:conf/iclr/BharadhwajBEL22} for AS distractors), and we follow this practice in \cref{section_experiment_performance}. To further solidify the effectiveness of our method, we present more experimental results on a broader range of environments: (1) Car Racing from OpenAI Gym \cite{brockman2016openai}, a challenging control task from pixels known for sudden road changes in the environment, which is used in \cite{ha2018world, rafailov2021visual, huang2022action}, and (2) Jaco Arm (Reach Top Left), a 6-DOF robotic arm task utilized in \cite{laskin2021urlb}. We cut out the bottom part of the image frames in Car Racing to prevent the agent from directly learning from the reward signals. Then, we augment these tasks with heterogeneous or homogeneous visual distractors same as our practice in \cref{sec: Environments and Tasks} and \cref{appendix: Environments and Tasks}, resulting in three new tasks: Car Racing + NBV, Jaco Arm + NBV, and Jaco Arm + AS. We select TIA and Dreamer as baseline methods, since they represent the best baselines in NBV and AS settings, respectively, as is shown in \cref{fig: performance}. 

\textbf{Remarks} Incorporating shifted agents into the Car Racing environment is exceptionally challenging, hence we exclusively make evaluation in the NBV setting for this task. For Jaco Arm + AS, we introduce an additional robotic arm into the observation, with the same connection point to the machine as the original arm. This simulates a scenario where the robot appears to have two arms for control and reaching the goal, and the policy needs to distinguish between the two arms to identify the controllable one. The example observations of these new tasks are shown in \cref{fig: carjaco_3_envs}.

The results presented in \cref{tb: carjaco results} below (Mean and Std returns of final test performance by running 10 episodes over 4 seeds) indicate that AD3 continues to outperform other methods on these tasks. Notably, despite performing relatively well on DMC tasks as demonstrated in \cref{section_experiment_performance}, Dreamer falls significantly behind our AD3 method on JacoArm + AS. This substantial performance gap clearly underscores the superiority of our method when dealing with homogeneous distractors.

\begin{table}[h]
\centering
\caption{Performance on more visual control tasks with different types of visual distractors }
\vskip 0.1in
\begin{small}
\begin{sc}
\begin{tabular}{lccc}
    \toprule
    {Method} & {Car Racing + NBV} & {Jaco Arm + NBV} & {Jaco Arm + AS}\\
    \midrule
    Dreamer & 251 $\pm$ 129 & 3 $\pm$ 8 & 7 $\pm$ 20 \\
    TIA & 304 $\pm$ 177 & 3 $\pm$ 9 & 35 $\pm$ 51 \\
    AD3 (Ours) & \textbf{505 $\pm$ 174} & \textbf{10 $\pm$ 25} & \textbf{58 $\pm$ 72}  \\
    \bottomrule
\end{tabular}
\end{sc}
\label{tb: carjaco results}
\end{small}
\end{table}

\subsection{Could Pretraining Benefits the Performance of AD3 on Extremely Hard Tasks?} 

\begin{figure*}[tbp]
\begin{center}
\centerline{\includegraphics[width=0.6\textwidth]{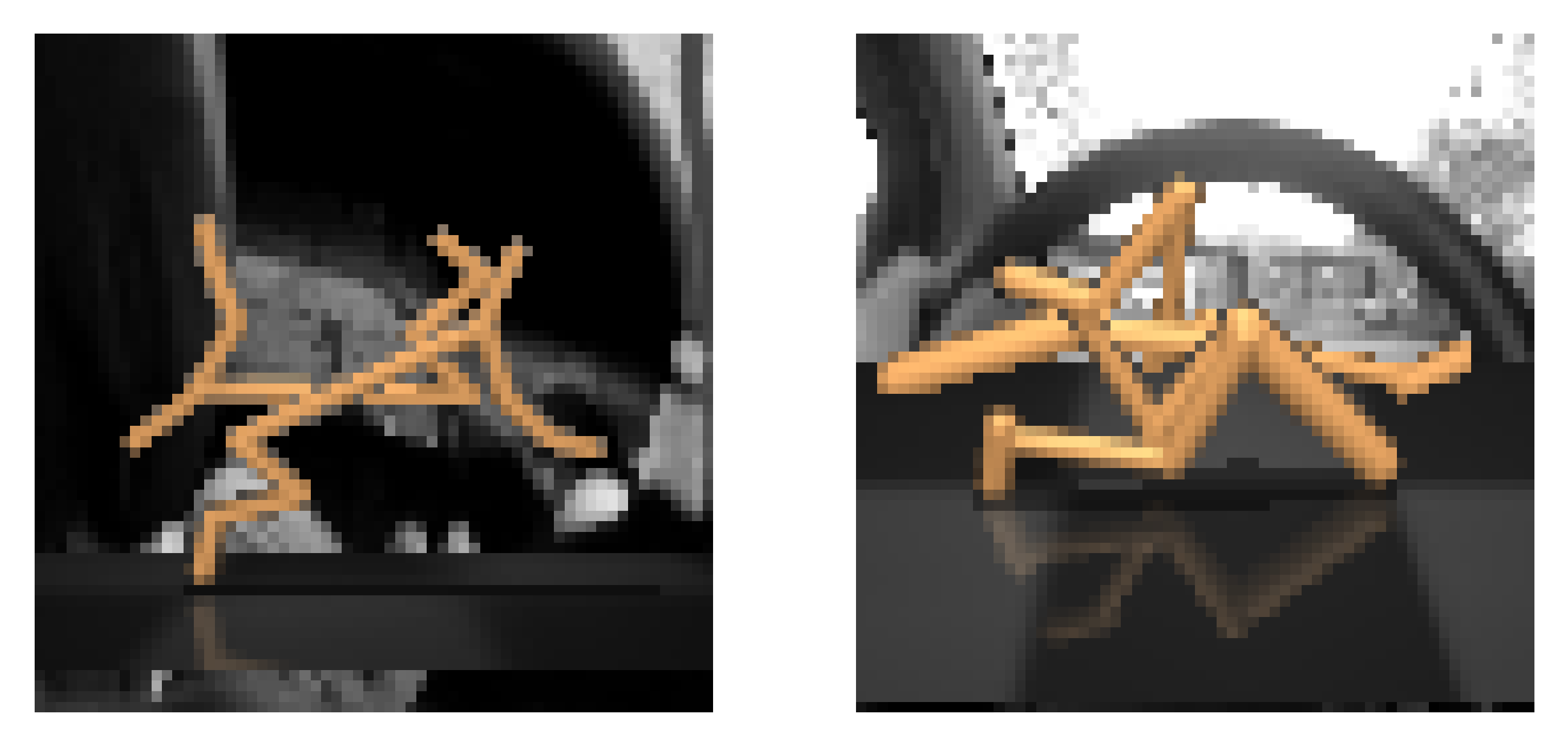}}
\vskip -0.1in
\caption{The example observations of the environments featuring \textit{Cheetah Run} and \textit{Walker Walk}, both equipped with NBV + AS distractors. These challenging tasks pose significant difficulties in inferring the implicit actions of the complex distractors.}
\label{fig: NBVAS_2_envs}
\end{center}
\vskip -0.2in
\end{figure*}

Modeling the distractors in real-world applications from more complex dynamics and visual inputs is indeed challenging. Nevertheless, we propose several potential ways to scale AD3 to more difficult tasks. Firstly, pretraining IAG on offline datasets to model the motion of underlying distractors can facilitate online finetuning. Furthermore, if the distractors have explicitly separated semantics, partitioning the implicit action $a^-$ into non-overlapping parts and incrementally training different parts of $a^-$ can be effective. For instance, in phase 1: $(o_t, o_{t+1}, a_t^+) \rightarrow a_t^{-^{(1)}}$, and in phase 2: $(o_t, o_{t+1}, a_t^+, a_t^{-^{(1)}}) \rightarrow a_t^{-^{(2)}}$, and so forth.

Apparently, an even harder visual distractor can be achieved by combining the above two mentioned backgrounds, which is called Agent Shifted with Natural Video Backgrounds (NBV + AS) that combines Natural Video Background (NBV) and Agent Shifted (AS) distractors, which is also based on the InfoPower work \cite{DBLP:conf/iclr/BharadhwajBEL22}. To validate our proposed approach, we instantiate scenarios such as \textit{Cheetah Run} + NBV + AS and \textit{Walker Walk} + NBV + AS, where inferring the implicit actions of homogeneous distractors becomes more difficult given the complex NBV dynamics, and vice versa. The example observations of these tasks are shown in \cref{fig: NBVAS_2_envs}.

We propose a method based on incremental pretraining-finetuning to implement the approaches mentioned above: first, pretrain only the IAG module offline on pre-collected datasets with NBV distractors, and then perform online finetuning of the entire AD3 on corresponding NBV + AS environments. In online incremental finetuning, we utilize a new IAG to infer implicit actions of shifted agents, conditioned on implicit actions of NBV inferred by the frozen pretrained IAG. Naturally, another available method involves swapping the order of AS and NBV. We compare these incremental methods to the original AD3 (directly training on NBV + AS environments). The results are exhibited in \cref{tb: NBVAS results}.

\begin{table}[h]
\centering
\caption{Performance on extremely hard tasks with NBV + AS distractors }
\vskip 0.1in
\begin{small}
\begin{sc}
\begin{tabular}{lcc}
    \toprule
    {Method} & {Cheetah Run + NBV + AS} & {Walker Walk + NBV + AS}\\
    \midrule
    AD3 & 422 (500K), \textbf{642 (1M)} & 486 (500K), 532 (1M) \\
    AD3\_Incremental\_NBV$\rightarrow$AS & 538 (500K) & \textbf{748 (500K)} \\
    AD3\_Incremental\_AS$\rightarrow$NBV & 534 (500K) & \textbf{771 (500K)}  \\
    \bottomrule
\end{tabular}
\end{sc}
\label{tb: NBVAS results}
\end{small}
\vskip -0.05in
\end{table}

For incremental methods, we perform online finetuning for 500K steps after pretraining. In contrast, the baseline AD3 is directly trained for 1M steps. The results demonstrate that offline pretraining significantly facilitates online finetuning, as incremental methods outperform AD3 with the same number of online learning steps (500K), aided by pretraining. They even match or surpass  the performance of the oracle method AD3 (1M). This underscores the benefit of incremental methods in inferring implicit actions of more complex distractors in more challenging environments. Furthermore, the two incremental methods yield similar results, showcasing the robustness and scalability of IAG.

\section{Additional Discussions}

\subsection{Assumptions Regarding State Decomposition in Visual RL}
\label{appendix: assumptions decomposition}
In real-world scenarios, it is common for latent states to be decoupled based on task relevance. For example, consider book-finding robots in libraries that scan shelves to locate specific books. These robots encounter both task-relevant objects (e.g., books) and task-irrelevant objects (e.g., background posters or people moving by), which need to be separated and distinguished by the robot.

Prior research has frequently made assumptions of decoupling different aspects of the latent state in terms of controllability or reward relevance when addressing visual distractors. TIA \cite{DBLP:conf/icml/FuYAJ21} assumes the existence of task-relevant and irrelevant parts, with both having their dynamics conditioned on agent actions. However, agent action inherently carries limited information about task-irrelevant dynamics of the background, which makes it a must for TIA to add additional designs and constraints (i.e., the extra loss mentioned in \cref{sec: Introduction}) and in turn presents learning challenges. EX-BMDP \cite{efroni2021provable}, Iso-Dream \cite{pan2022iso} and SeMAIL \cite{wan2023semail} modify the TIA assumption by removing the task-irrelevant part's dependence on any action, treating it as an uncontrollable component. Although sounds intuitive, these methods still require extra efforts to prevent the task component from capturing all the information, and may fail to distinguish the homogeneous distractors. Denoised MDP \cite{DBLP:conf/icml/0001D0IZT22} learns world models for different components factorized by reward and action according to the assumed transition structures. However, the actions employed for learning these MDPs may not suffice to effectively disentangle the various components, particularly when homogeneous distractors are encountered, as is empirically demonstrated in \cref{section_experiment_performance}. Our IABMDP assumption deviates from prior methods primarily by explicitly introducing the concept of distractor actions to extract task-irrelevant components. This distinctive feature sets our method apart from existing approaches.

Furthermore, there are instances where an initially irrelevant distractor may become task-relevant under certain conditions, thereby influencing the agent's decision-making process. In such cases, we propose incorporating a brief period of the particular distractor states into the policy function during training. This approach enables the agent to adapt to changing conditions and make informed decisions based on relevant information.

\subsection{Computational Costs}
The training of the Implicit Action Generator (IAG) may add computational overhead to the modeling process. However, the introduction of these additional costs is justified by the substantial performance gains achieved over TIA and Dreamer. The use of implicit actions inferred by IAG facilitates the construction of separated world models, leading to improved performance. 

Moreover, the extra computational costs are limited to the training phase. During execution, the IAG module is not needed to infer implicit actions. Instead, we utilize the trained task-relevant world model to encode observations into latent space and generate action outputs using the policy, which incurs similar computational costs as Dreamer and TIA. Furthermore, by eliminating two loss objectives (adversarial reward dissociation and distractor-only image reconstruction) from TIA, we in turn effectively reduce computational costs. 

We acknowledge the potential for exploring more efficient network designs and training methods for applying IAG in real-world scenarios, and we will address these aspects in future work.

\subsection{Weakness of TIA When Dealing with Homogeneous Distractors}
\label{appendix: Weakness of TIA}
TIA seeks to disentangle the latent state into task-relevant and task-irrelevant components, using adversarial reward dissociation (RD) on the task-irrelevant state and distractor-only image reconstruction (DOR) by the task-irrelevant component. However, these loss objectives are highly sensitive, as the optimization of RD can be unstable, and finding the right balance between RD and DOR is challenging. Without proper tuning of the weights for these two objectives, TIA may inadvertently invert the two components. Additionally, TIA utilizes agent actions to model the dynamics of distractors, which is inappropriate and can further contribute to the inversion of these components.

Furthermore, the assumption underlying DOR may not hold in environments with homogeneous distractors. DOR is introduced to prevent the task model from reconstructing the entire observation by itself and containing too much information, so TIA incorporates an additional image decoder that encourages the distractor model to reconstruct the entire observation by itself. However, there is an implicit assumption behind DOR: the task-relevant information comprises a relatively small proportion of the observation (e.g., in NBV distractors commonly used in prior work, where noisy videos dominate most of the observation space). When confronted with homogeneous distractors (e.g., shifted agents), task-relevant and task-irrelevant information occupy nearly equal portions of the observation space. Encouraging the distractor state to reconstruct the entire image may lead to an overly dominant distractor state, potentially causing a reversal of the two components once again.

We strongly recommend the reader to refer to \href{https://sites.google.com/view/ad3-iag}{https://sites.google.com/view/ad3-iag} for video visualizations demonstrating TIA's failure in tackling homogeneous distractors.

\subsection{Categorical Latents}
\label{appendix: categorical latents}
We utilize categorical variables to bottleneck the implicit action space in IAG, thereby avoiding shortcuts. Each implicit action is represented by a vector of multiple categorical (one-hot) variables. In our ablation study (\cref{sec: ablation study}), categorical variables significantly outperform other bottlenecking methods such as Vector Quantization (VQ). Compared to traditional quantization methods, categorical (one-hot) coding is more flexible because the learning process of the codebook and the quantizing operation are inherently embedded in the categorical-variable architecture. Using one-hot codes in forward propagation equals to leveraging active bits in the code as indices for quantization. Consequently, different latent codebooks can be implicitly built for different types of loss objectives (e.g., in the decoder of cycle consistency, difference reconstruction, and one-step reconstruction), resulting in improved learning performance. Additionally, the size of the latent codebook and the dimension of the vector representation do not need to be specified in advance, and the burden of optimizing the codebook loss and the commitment loss is also eliminated.

\begin{figure}[H]
\begin{center}
\centerline{\includegraphics[width=0.85\textwidth]{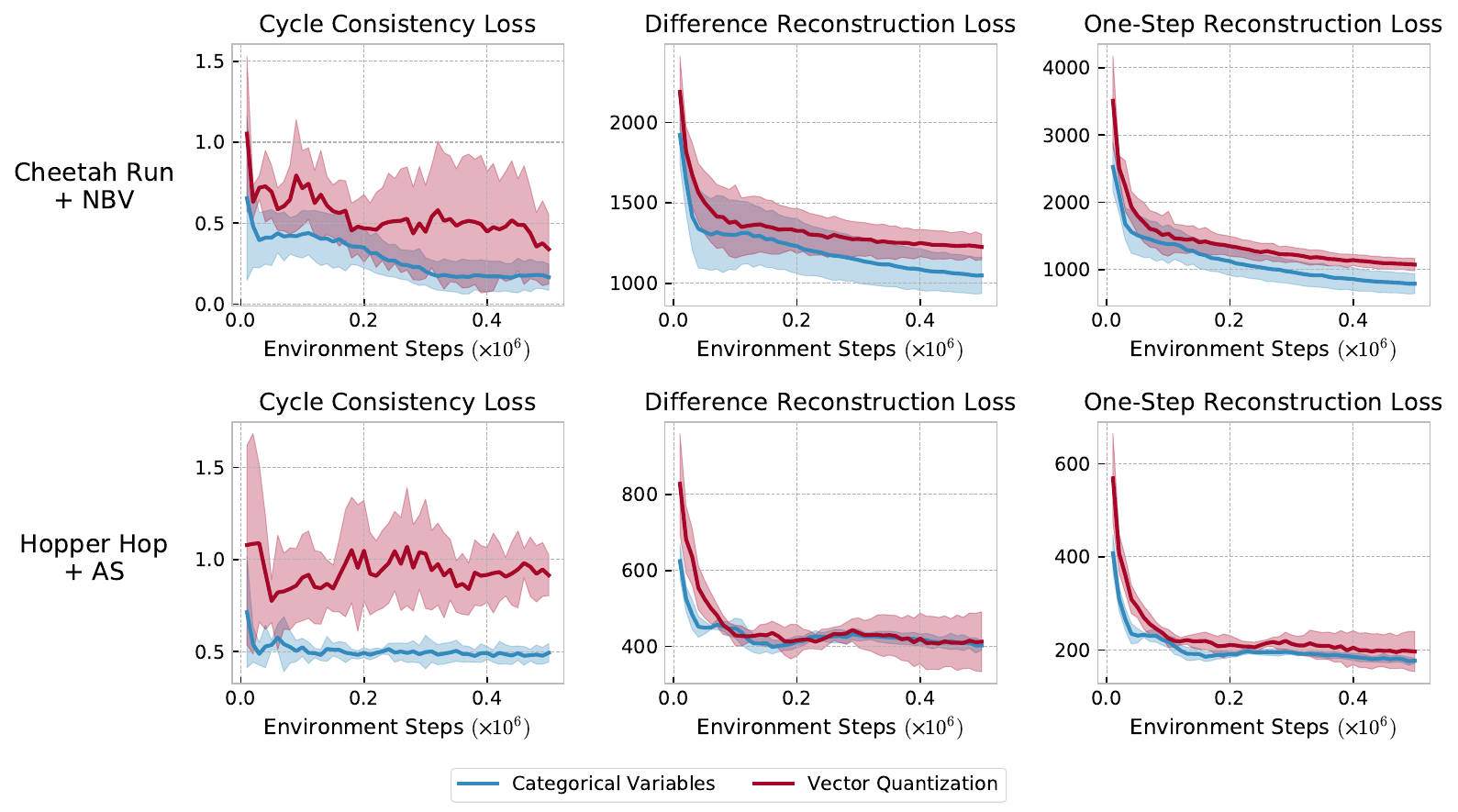}}
\vskip -0.05in
\caption{Loss curves of the three learning objectives in IAG on two tasks with different distractors: \textit{Cheetah Run} + NBV, and \textit{Hopper Hop} + AS. Each experiment is conducted over 4 seeds. The solid curves and the shaded region represent the average loss value and the standard error across different runs, respectively. All three loss functions exhibit a more rapid decrease and reach comparable or lower values on both tasks when categorical variables are employed, particularly for the cycle consistency loss.}
\label{fig: ohq_vq_three_loss_curves}
\end{center}
\vskip -0.2in
\end{figure}

We present the loss curves for the three learning objectives of IAG in \cref{eq:IAG} on two tasks, comparing the use of categorical variables and VQ in \cref{fig: ohq_vq_three_loss_curves}. All three loss functions exhibit a more rapid decrease and reach comparable or lower values on both tasks when categorical variables are employed, particularly for the cycle consistency loss. Therefore, IAG learned with categorical variables performs significantly better in distinguishing task-irrelevant distractors than VQ in the AD3 algorithm, as demonstrated in our ablation study (\cref{sec: ablation study}). These results suggest that categorical variables provide a more efficient and robust representation for bottlenecking the latent space compared to vector quantization, especially in tasks involving complex environmental dynamics.

In \cref{fig: ohq_vq_three_loss_curves}, the gap between the final values of the two reconstruction loss for categorical variables and VQ in \textit{Hopper Hop} + AS is relatively smaller than in \textit{Cheetah Run} + NBV. This is because the precise reconstruction of complex noisy video backgrounds is more challenging than that of shifted agent backgrounds. However, this trend is reversed for the cycle consistency loss, whose final value in \textit{Hopper Hop} + AS is slightly higher, and the gap between the two representations is larger compared to \textit{Cheetah Run} + NBV. As cycle consistency loss serves as a more critical learning objective than the other two losses (\cref{sec: ablation study}), this may indicate that distinguishing homogeneous distractors is more challenging than distinguishing heterogeneous ones, since heterogeneous distractors possess more visually distinguishable factors. Although precise modeling of NBV backgrounds is significantly more difficult, identifying which agent is controllable is more challenging in tasks with AS distractors. Furthermore, the larger gap in cycle consistency loss between the two representations in solving AS distractors compared to NBV ones further underscores the superiority of using categorical variables in identifying task-irrelevant dynamics.

In addition to its flexibility and effectiveness in optimizing different types of objective functions, the extreme sparsity enforced by categorical variables significantly benefits the bottlenecking of the implicit action space in IAG, preventing the action space from containing an excessive amount of information. Such a level of sparsity can also be beneficial for generalization. \cite{DBLP:conf/iclr/HafnerL0B21} highlights that categorical variables can also be a better inductive bias for modeling multi-modal changes between image frames and the non-smooth aspects of the environment, such as sudden transitions between looping background videos in the NBV environments. This property significantly benefits our implicit action model in capturing the complex semantics of environmental transitions. Further exploration is needed to understand the additional benefits of using categorical variables as discrete encodings and their applications in other areas.

\section{Additional Visualization Results}
\label{sec: Additional Visualization Result}
We provide visualization results of task-relevant and irrelevant reconstructions here. Moreover, we strongly recommend the reader to refer to \href{https://sites.google.com/view/ad3-iag}{https://sites.google.com/view/ad3-iag} for more extensive dynamic visualization results (videos for the reconstruction of imagined trajectories in task-relevant and irrelevant world models). These resources aim to offer a more intuitive understanding of AD3's proficiency in distinguishing both heterogeneous and homogeneous visual distractors.

For each image, from top to bottom: raw observation, reconstruction in the task-relevant / irrelevant latent state space, difference; from left to right: different image frames in 6 different trajectories. 

\begin{figure}[H]
\begin{center}
\includegraphics[width=0.9\textwidth]{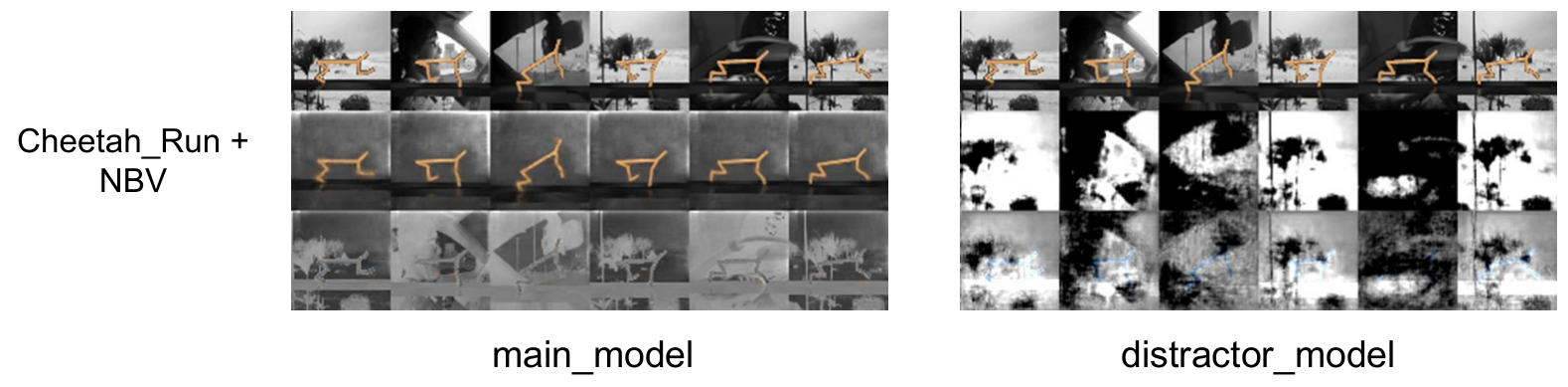}
\end{center}
\end{figure}

\begin{figure}[H]
\begin{center}
\includegraphics[width=0.9\textwidth]{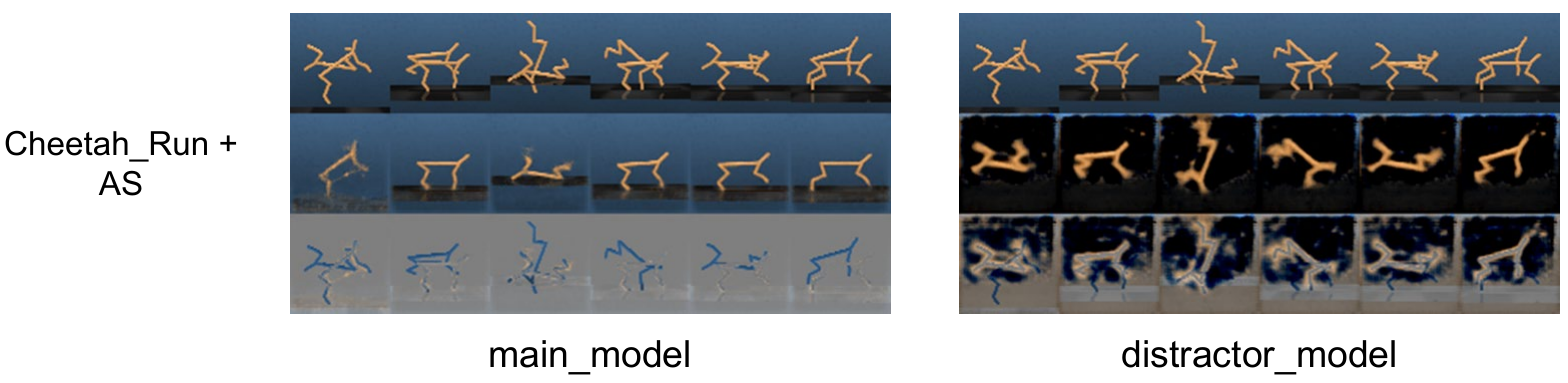}
\end{center}
\end{figure}

\begin{figure}[H]
\begin{center}
\includegraphics[width=0.9\textwidth]{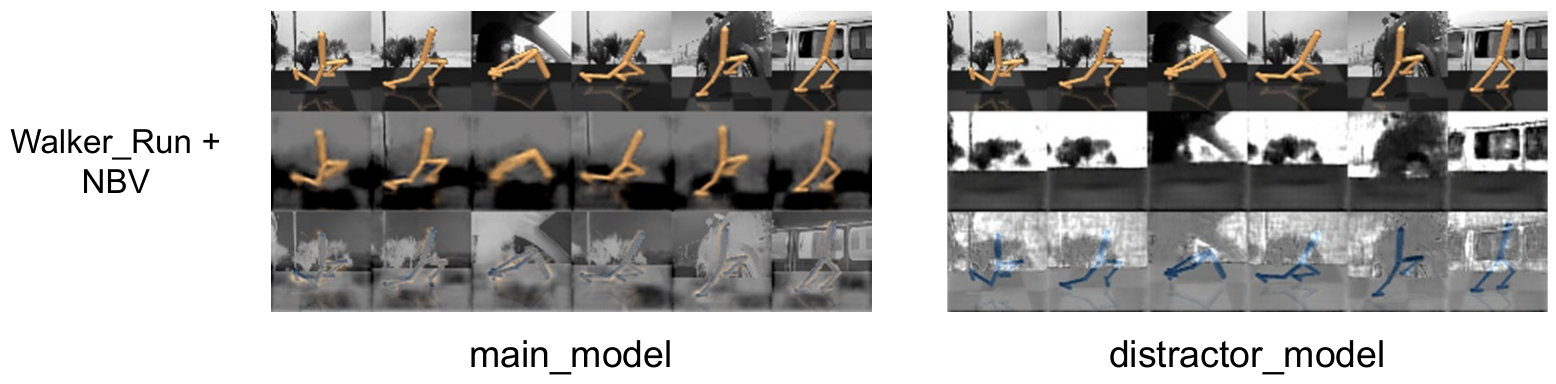}
\end{center}
\end{figure}

\begin{figure}[H]
\begin{center}
\includegraphics[width=0.9\textwidth]{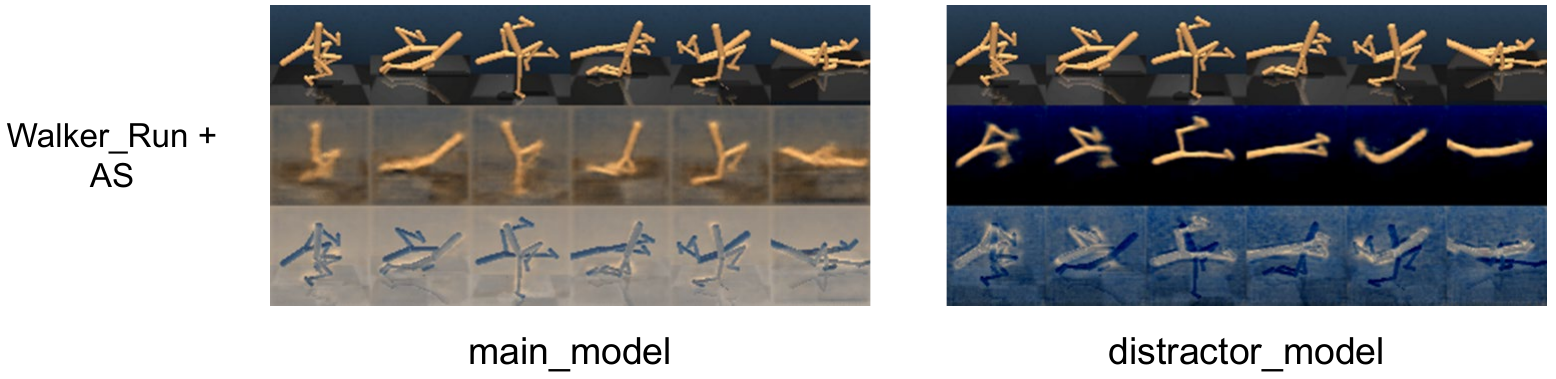}
\end{center}
\end{figure}

\begin{figure}[H]
\begin{center}
\includegraphics[width=0.9\textwidth]{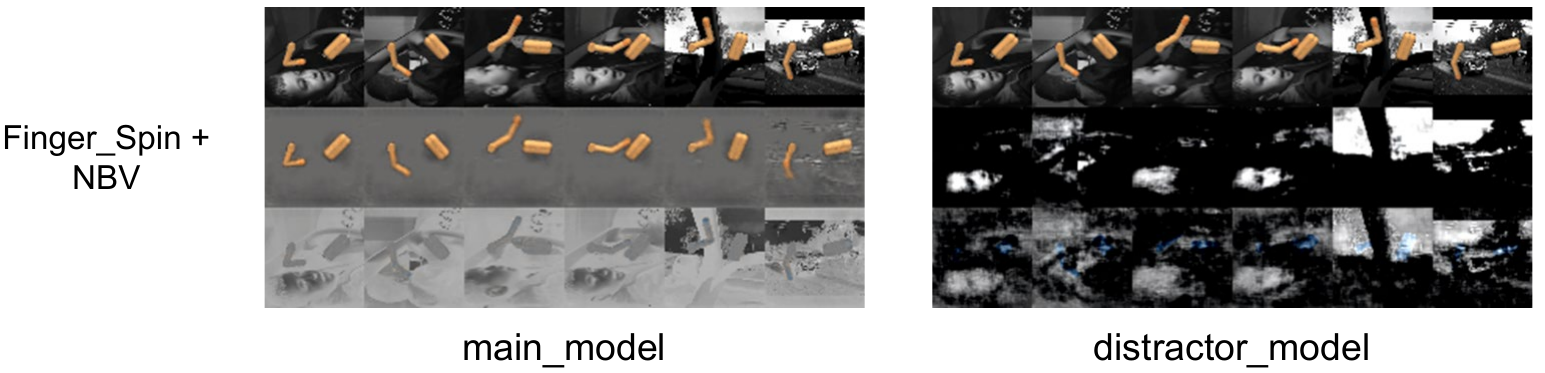}
\end{center}
\end{figure}

\begin{figure}[H]
\begin{center}
\includegraphics[width=0.9\textwidth]{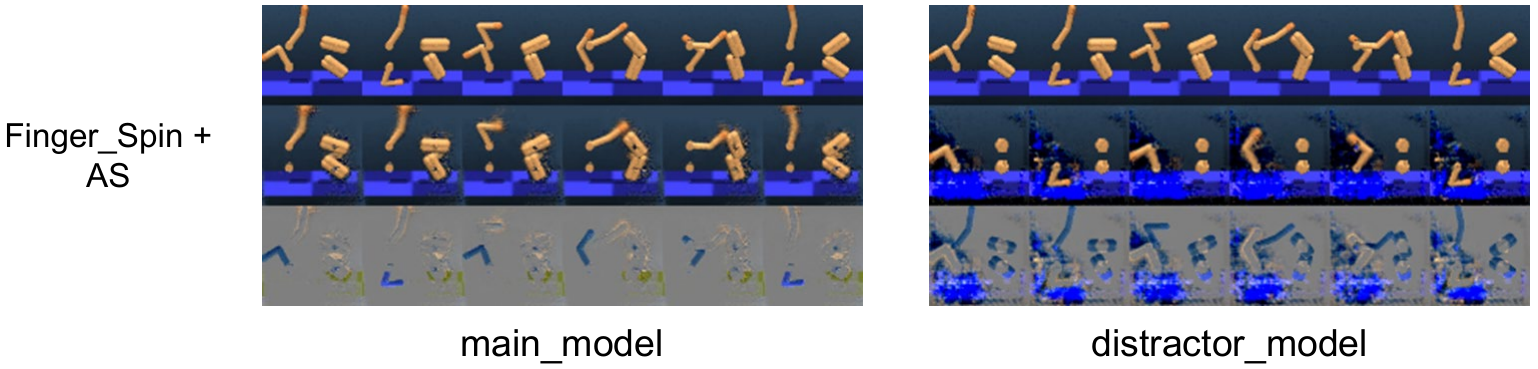}
\end{center}
\end{figure}

\begin{figure}[H]
\begin{center}
\includegraphics[width=0.9\textwidth]{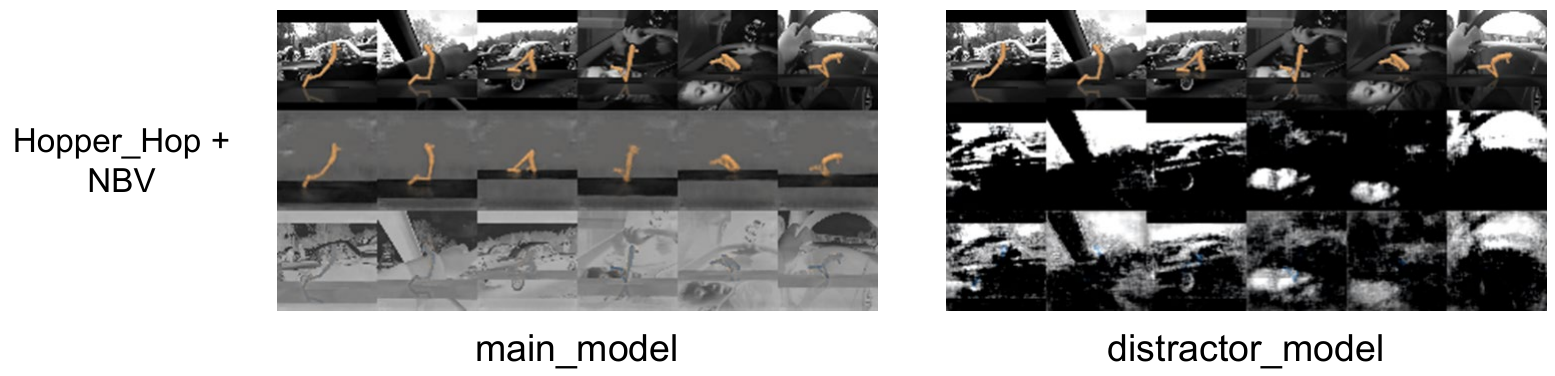}
\end{center}
\end{figure}

\begin{figure}[H]
\begin{center}
\includegraphics[width=0.9\textwidth]{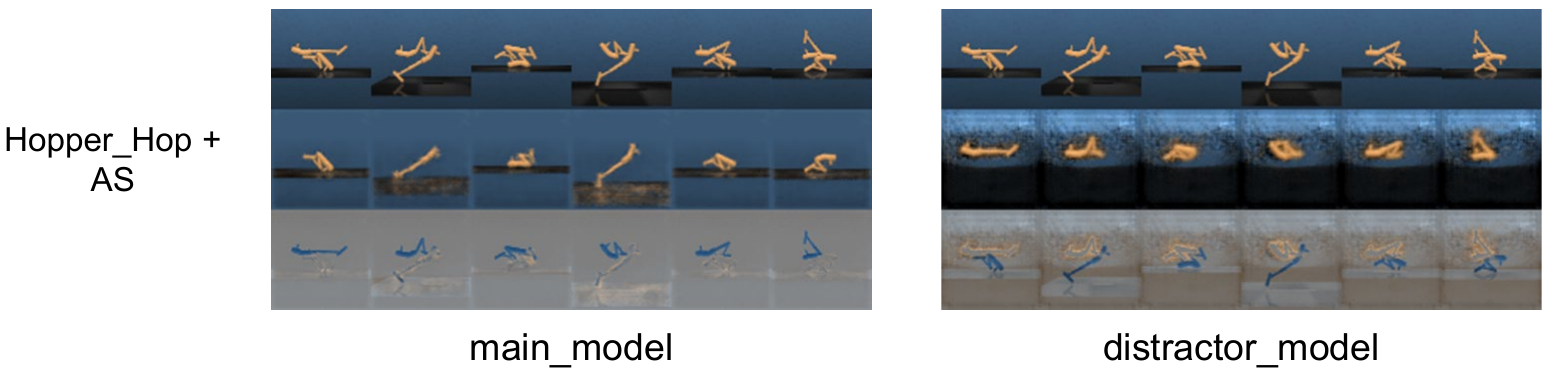}
\end{center}
\end{figure}

\end{document}